\documentclass[10pt,twocolumn,letterpaper]{article}

\usepackage{cvpr}              

\usepackage[accsupp]{axessibility}  
\definecolor{cvprblue}{rgb}{0.21,0.49,0.74}
\usepackage[pagebackref,breaklinks,colorlinks,allcolors=cvprblue]{hyperref}
\usepackage{gradient-text}
\usepackage{xcolor}
\usepackage{pifont}
\usepackage{bbding}
\usepackage{multirow}
\usepackage{tikz}
\usepackage{graphicx}
\usepackage{subcaption}
\usepackage{amsmath, amssymb, amsthm}
\usepackage{fdsymbol}
\usetikzlibrary{arrows.meta}
\definecolor{deep_purple}{RGB}{120,44,180}

\usepackage{colortbl}

\definecolor{newgreen}{rgb}{0.7, 0.9, 0.7}
\definecolor{newblue}{rgb}{0.85, 0.85, 0.9}
\definecolor{tkcolor}{RGB}{224,223,255}
\definecolor{shallowred}{RGB}{242,204,208}
\definecolor{shallowgray}{RGB}{237,240,246}

\usepackage{todonotes}
\setuptodonotes{inline, bordercolor=white, backgroundcolor=tkcolor}

\usepackage{booktabs}

\definecolor{spadecolor}{rgb}{1 , 0.7 , 0}
\definecolor{clubcolor}{rgb}{0.9 , 0.3 , 0.4}
\definecolor{heartcolor}{rgb}{0.8 , 0.4 , 0.9}
\definecolor{diamondcolor}{rgb}{0. , 0.5 , 0.9}
\definecolor{circlecolor}{rgb}{0.3 , 0.9 , 0.7}
\definecolor{squarecolor}{rgb}{0.5, 0.5, 0.5}

\newcommand{\clubsymbol}{\textcolor{clubcolor}{$\clubsuit$}}
\newcommand{\diamondsymbol}{\textcolor{diamondcolor}{$\vardiamondsuit$}}

\definecolor{upgreen}{HTML}{32CD32}
\definecolor{downred}{HTML}{DC143C}
\newcommand{\up}[1]{\textcolor{nicegreen}{\textbf{$+$#1}}}
\newcommand{\down}[1]{\textcolor{clubcolor}{\textbf{$-$#1}}}

\newcommand{\drawSimpleAxes}[2]{
\begin{tikzpicture}
    \draw[-{Stealth[length=4pt]},thick] (-0.1,0) -- (#1,0);
    \draw[-{Stealth[length=4pt]},thick] (0,-0.1) -- (0,#2);
\end{tikzpicture}
}
\newcommand{\drawThreeArrows}[1][1.0]{
\begin{tikzpicture}[scale=#1]
    \filldraw (0,0) circle (3pt);
    \draw[-{Stealth[length=4pt]},thick] (0,0) -- (1,0) coordinate (a);
    \draw[-{Stealth[length=4pt]},thick] (0,0) -- ({cos(30)}, {sin(30)}) coordinate (b);
    \draw[-{Stealth[length=4pt]},thick] (0,0) -- ({cos(60)}, {sin(60)}) coordinate (c);
    \path (a) -- ++(0,{sin(30)-sin(60)/2}) coordinate (d);
    \path (b) -- ++(0,{sin(60)-sin(30)/2}) coordinate (e);
    \path (c) -- ++(0,{sin(30)-sin(60)/2}) coordinate (f);
\end{tikzpicture}
}

\usepackage{stackengine}
\DeclareRobustCommand{\wuline}[1]{%
  \setbox0=\hbox{#1}%
  \stackengine{-1pt}{#1}{\smash{\raisebox{-0.3ex}{%
    \hspace{-0.1em}%
    \makebox[\wd0][l]{\xleaders\hbox{\texttildelow\kern-.32em}\hfill}%
  }}}{O}{c}{F}{F}{L}%
}

\usepackage{tcolorbox}
\definecolor{shallowgray}{RGB}{237,240,246}
\definecolor{tldrlightpurple}{RGB}{242,242,255}
\newtcolorbox{prompt}[1][]{
    width=\columnwidth,
    colback = tldrlightpurple, 
    colframe = tldrlightpurple, 
    boxsep=0pt,left=10pt,right=10pt,top=5pt,bottom=5pt,
    fontupper=\small\linespread{0.9}\selectfont,
    title=#1
}

\newtcolorbox{promptwide}[1][]{
  enhanced,
  float*=t,
  width=\textwidth,
  colback = tldrlightpurple,
  colframe = tldrlightpurple,
  boxsep=0pt,left=10pt,right=10pt,top=5pt,bottom=5pt,
  fontupper=\small\linespread{0.9}\selectfont,
  title=#1
}

\definecolor{takeawaybg}{HTML}{f3f8fb}    
\definecolor{takeawayborder}{HTML}{6faad0} 
\definecolor{takeawaytitle}{HTML}{bcd7e9}  
\definecolor{takeawayshadow}{HTML}{d8d8d8} 

\newtcbox{\highlighttext}{%
    on line,
    tcbox width=auto limited,
    tcbox raise base,
    nobeforeafter,
    colback=takeawaybg,
    colframe=takeawayborder,
    boxrule=1pt,
    arc=2pt,
    boxsep=1pt,
    left=2pt, right=2pt,
    top=1pt, bottom=1pt,
    after=\raggedright
}

\newcommand\encircle[2][]{\tikz[overlay]\node[fill=blue!20,inner sep=2pt, anchor=text, rectangle, rounded corners=1.5mm,#1] {#2};\phantom{#2}}
\definecolor{lightcoral}{rgb}{0.94, 0.5, 0.5}
\definecolor{myblue}{rgb}{0.27,0.52,0.95}
\definecolor{harvestgold}{rgb}{0.85, 0.57, 0.0}

\definecolor{nicegreen}{RGB}{34,174,24}

\newcommand{\cmark}{{\color{nicegreen} \ding{51}}}
\newcommand{\xmark}{{\color{clubcolor} \ding{55}}}

\definecolor{navyblue}{HTML}{0071BC}
\definecolor{hotpink}{HTML}{FF0080}

\definecolor{backblue}{RGB}{210,230,250}

\definecolor{backred}{RGB}{255,203,203}

\definecolor{backgreen}{RGB}{190,240,210}

\usepackage{multirow}
\usepackage{booktabs}
\usepackage{makecell}

\usepackage{fontawesome5}

\usepackage{marvosym} 

\def\paperID{11221} 
\def\confName{CVPR}
\def\confYear{2026}

\title{
\raisebox{-0.25em}{\includegraphics[height=1.45em]{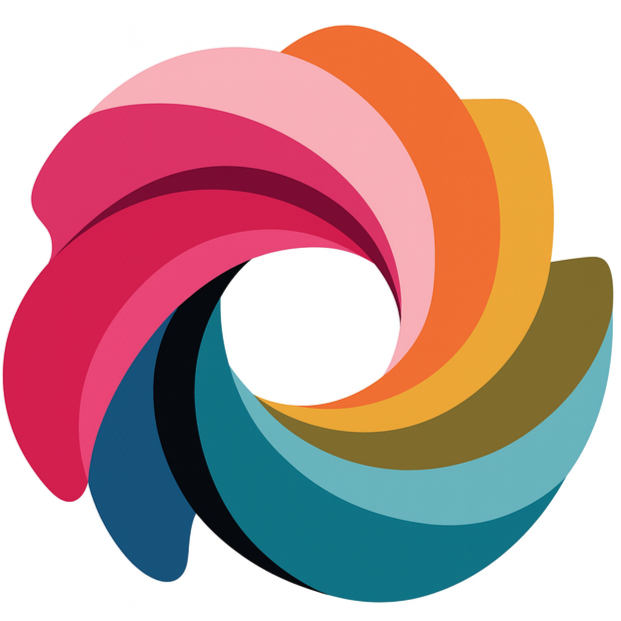}} \gradientRGB{Think360}{255,76,90}{120,44,180}\textcolor{deep_purple}{\textbf{$^\circ$}}

Beyond Depth: Evaluating the Width-centric Reasoning Capability of MLLMs}

\author{
  Mingrui Chen$^{1,2,4}$ Hexiong Yang$^{2,3}$ Haogeng Liu$^{1,2}$ Huaibo Huang$^{1,2,\text{\Letter}}$ Ran He$^{1,2,4}$ \\
  $^1$School of Artificial Intelligence, University of Chinese Academy of Sciences \\
  $^2$NLPR\&MAIS, Institute of Automation, Chinese Academy of Sciences \\
  $^3$School of Advanced Interdisciplinary Science, University of Chinese Academy of Sciences \\
  $^4$Zhongguancun Academy, $^{\text{\Letter}}$ Corresponding Authors \\
  \texttt{charmier2003@gmail.com, huaibo.huang@cripac.ia.ac.cn, rhe@nlpr.ia.ac.cn} \\
}

\begin{document}
\maketitle
\begin{abstract}
In this paper, we present a holistic multimodal benchmark that evaluates the reasoning capabilities of MLLMs with an explicit focus on reasoning \texttt{\textbf{width}}, a complementary dimension to the more commonly studied reasoning \texttt{\textbf{depth}}.
Specifically, reasoning depth measures the model’s ability to carry out long-chain, sequential reasoning in which each step is tightly and rigorously linked to the next.
Reasoning width tends to focus more on the model’s capacity for broad trial-and-error search or multi-constrained optimization: it must systematically traverse many possible and parallelized reasoning paths, apply diverse constraints to prune unpromising branches, and identify valid solution routes for efficient iteration or backtracking. 
To achieve it, we carefully curate 1200+ high‑quality multimodal cases spanning heterogeneous domains, and propose a fine-grained tree-of-thought evaluation protocol that jointly quantifies reasoning \textit{width} and \textit{depth}. 
We evaluate \textbf{12} major model families (over \textbf{30} advanced MLLMs) across difficulty tiers, question types, and required skills. 
Results show that while current models exhibit strong performance on general or common-sense VQA tasks, they still struggle to combine deep sequential thought chains with wide exploratory search to perform genuine insight-based reasoning. Finally, we analyze characteristic failure modes to provide possible directions for building MLLMs that reason not only \textit{deeper} but also \textit{wider}. Code will be available at \href{https://github.com/Walnutes/Think360}{Think360}.
\end{abstract}

\section{Introduction}
Over the past decade, deep learning has advanced at a rocket-like pace, driven by an ever-tightening interplay between \textit{\textbf{models}}\footnote{including novel neural architectures designs, training paradigms, and scaling recipes.}~\cite{fan2024rmt, dosovitskiy2020image, he2022masked, muennighoff2025s1, kaplan2020scaling} and \textit{\textbf{benchmarks}}~\cite{lu2023mathvista, yang2025had, zhang2024mathverse, wang2024measuring, phan2025humanity} that continually raise the bar. Echoing the \textit{Second Half of AI}, this dynamic forms a modern \texttt{spear-and-shield} contest: novel training techniques propel models to “hill-climb” existing leaderboards, while tougher benchmarks emerge in response, perpetuating the cycle of progress.

\begin{figure}[t]
    \centering
    \vspace{-0.5cm}
    \includegraphics[width=1\linewidth]{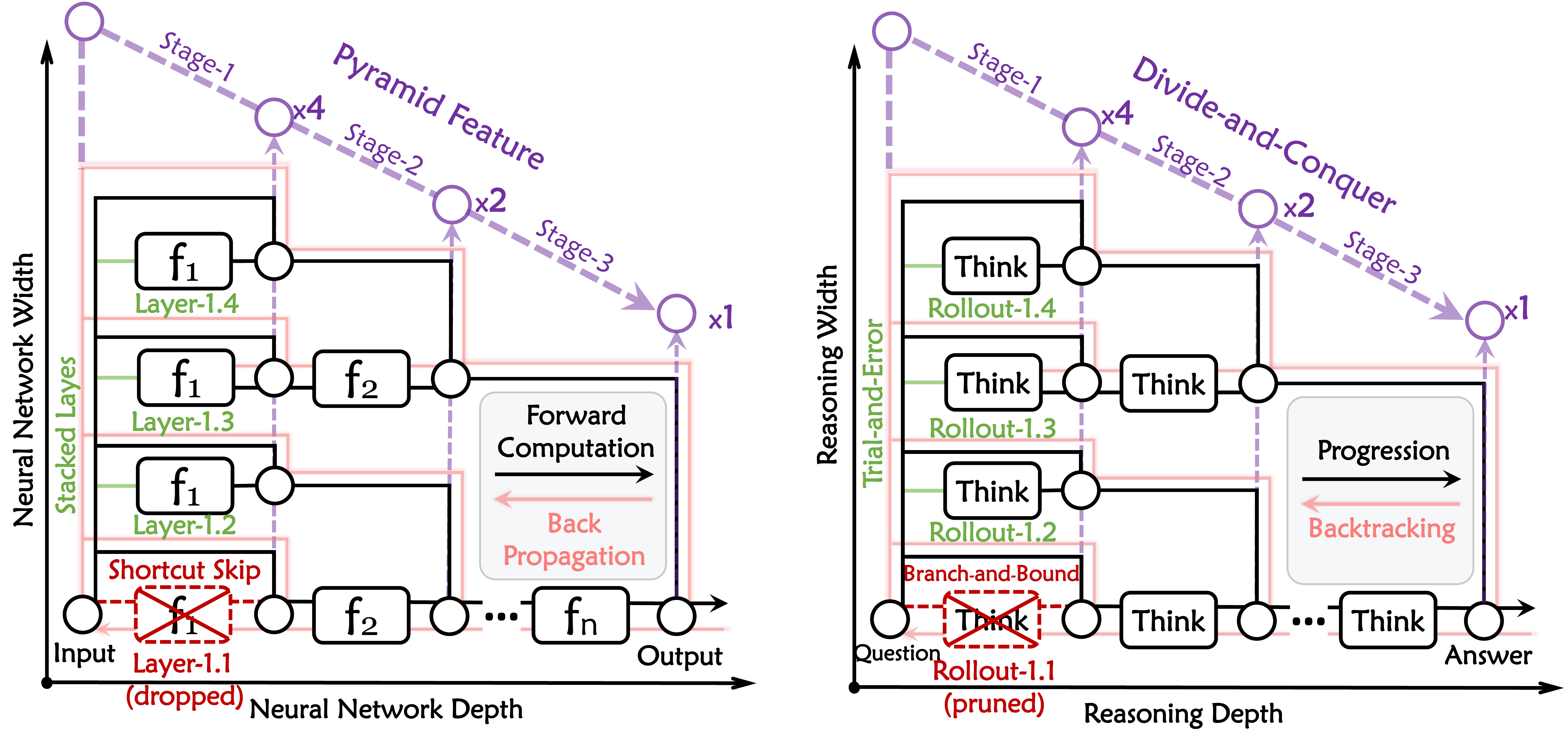}
    \caption{The concepts illustration for the \texttt{width} and \texttt{depth} in the information propagation process of neural network and reasoning. Drawing insights from the classical designs in neutral network: shortcut skipping or dropout, pyramid feature, layer stacking and gradient back propagation, we analogize these to the strategies: pruning, divide-and-conquer, trial-and-error and backtracking to distinguish depth versus width in inference processes.}
    \vspace{-0.5cm}
    \label{fig:concept}
\end{figure}

\begin{figure*}[ht]
    \centering
    \vspace{-0.9cm}
    \includegraphics[width=1\linewidth]{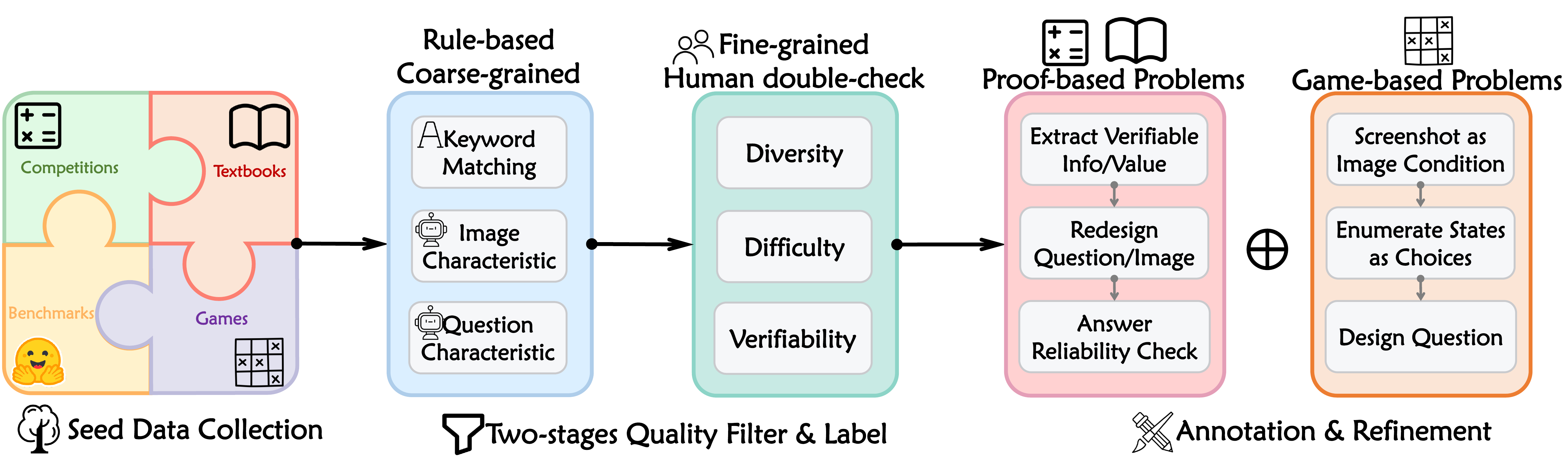}
    \caption{Three‑stage pipeline for constructing Think360$^\circ$—beginning with diverse seed data collection, progressing through a two‑step quality filter (rule‑based heuristics and human double‑check), and finalized through targeted annotation \& refinement (as demonstrated by proof‑ and game‑based problems).}
    \label{fig:workflow}
    \vspace{-0.4cm}
\end{figure*}

Nowhere is this loop more apparent than in the realm of large reasoning models (LRMs)~\cite{guo2025deepseek}. The promise of test-time scaling has inspired vigorous exploration of approaches ranging from training-free approach (\textit{e.g}., chain-of-thought prompting~\cite{wei2022chain}) to post-training methods (\textit{e.g}, supervised finetuning~\cite{li2024entropic} or reinforcement learning~\cite{shao2024deepseekmath} based preference alignment). Benchmarks have also evolved in lockstep, expanding (i) difficulty, from K12 level problems to graduate-level~\cite{rein2024gpqa} and even Olympiad-level~\cite{sun2025challenging} challenges; (ii) task coverage, from commonsense QA~\cite{antol2015vqa} to complex coding~\cite{jimenez2023swe} or more advanced mathematics~\cite{sun2025challenging}; and (iii) modality, from purely textual settings~\cite{rein2024gpqa} to richly multimodal inputs~\cite{zhang2024mathverse, lu2023mathvista, deng2025longdocurl} (and even multimodal outputs~\cite{hurst2024gpt}).

Yet most existing evaluations share one hidden axis: they tend to probe \textit{reasoning depth} (see Tab.\ref{tab:comparison_bench})—how far a model can extend a single \underline{reasoning chain}. Depth alone, however, paints an incomplete picture. Human problem-solvers rarely succeed by linear deduction alone; they multi-directionally navigate \raisebox{-0.5ex}{\drawSimpleAxes{0.2}{0.2}} the solution space $360$°, iteratively branching outward \raisebox{-0.5ex}{\drawThreeArrows[0.3]} from thought anchors across different conceptions, pruning $\nrightarrow$ dead-end paths, circling back $\circlearrowright$ to revisit alternative hypotheses or recombine partial trails until insight crystallizes and reaches the final answer.

To better understand this distinction, Fig.\ref{fig:concept} illustrates the correspondence between depth and width in neural networks and reasoning processes. In neural architectures, depth enables sequential feature abstraction through layered processing, while width manifests as parallel pathways that capture diverse representations—from multi-directional textures to frequency patterns. MLLMs reasoning similarly embodies this structure: trial-and-error when exploring multiple solution paths exemplifies breadth-first reasoning. Furthermore, neural design principles like shortcut connections, dropout, backpropagation, and pyramidal features parallel reasoning strategies in problem-solving: pruning eliminates dead-end branches, backtracking recovers from failures, and divide-and-conquer decomposes complex problems into manageable subcomponents. We further present a comparative analysis of the scaling paradigms of reasoning \texttt{depth} and \texttt{width} in \textcolor{red}{Appendix}.

Think360$^\circ$ is designed to assess whether models can \ding{182} systematically probe the solution space through systematic trial-and-error, \ding{183} juggle multiple simultaneous constraints to prune infeasible reasoning trajectory or pathway efficiently, and \ding{184} unify partial clue or discovery into final coherent answer—all while reasoning over both language and vision. By providing carefully curated tasks that demand expansive, \textbf{non-monotonic} exploration in addition to sole deep logical chains, Think360$^\circ$ tends to offer a comprehensive benchmark for width-centric multimodal reasoning.
\begin{figure*}[t]
\centering
\resizebox{\textwidth}{!}{
\begin{minipage}{0.47\textwidth} 
    \centering
    \vspace{-0.5cm}
    \setlength{\tabcolsep}{13pt}
    \begin{tabular}{lc}
         \toprule[1.5pt]
         \textbf{Statistic} & \textbf{Number} \\
         \midrule
          Total questions & 1225 (100\%)\\
          ~- \texttt{Testmini} set & 740 (60\%) \\
         \midrule
        Answer Type\\
          ~- Multiple-choice questions & 207 (16.9\%)\\
          ~- Free-form questions & 1018 (83.1\%)\\
          ~~~~\textbf{·} Numerical & 553 (54.3\%)\\
          ~~~~\textbf{·} Formula & 53 (5.2\%)\\
          ~~~~\textbf{·} Structure & 376 (37.0\%)\\
          ~~~~\textbf{·} Others & 36 (3.5\%)\\
         \midrule
         Difficulty Tier & 5 \\
         ~- Easy & 127 (10.4\%) \\
         ~- Basic & 272 (22.2\%) \\
         ~- Medium & 412 (33.6\%) \\
         ~- Hard & 293 (23.9\%) \\
         ~- Olympiad & 121 (9.9\%) \\
         \midrule
         Cognitive Capability/Skill$^*$ & 5 \\
         Question Type$^*$ & 6 \\
         \midrule
         Question Length \\
         ~- Maximum length & 369 \\
         ~- Average length & 82 \\

         Answer Length \\
         ~- Maximum length & 82 \\
         ~- Average length & 3 \\

         \bottomrule[1.5pt]
    \end{tabular}
 \captionof{table}{\textbf{Key statistics of Think360$^\circ$.} We construct a fine‑grained taxonomy across the dimensions of Difficulty Tier, and Answer Type, and additionally provide detailed statistics on question and answer lengths (unit of length: words). With regard to the non-exclusive categories marked with $^*$, we provide details in the following sections.}
 \vspace{-0.5cm}
 \label{tab:statics}
 \end{minipage}\hspace{0.01\textwidth}
 \begin{minipage}{0.46\textwidth}
 \centering
 \vspace{-0.5cm}
 \includegraphics[width=0.92\linewidth]{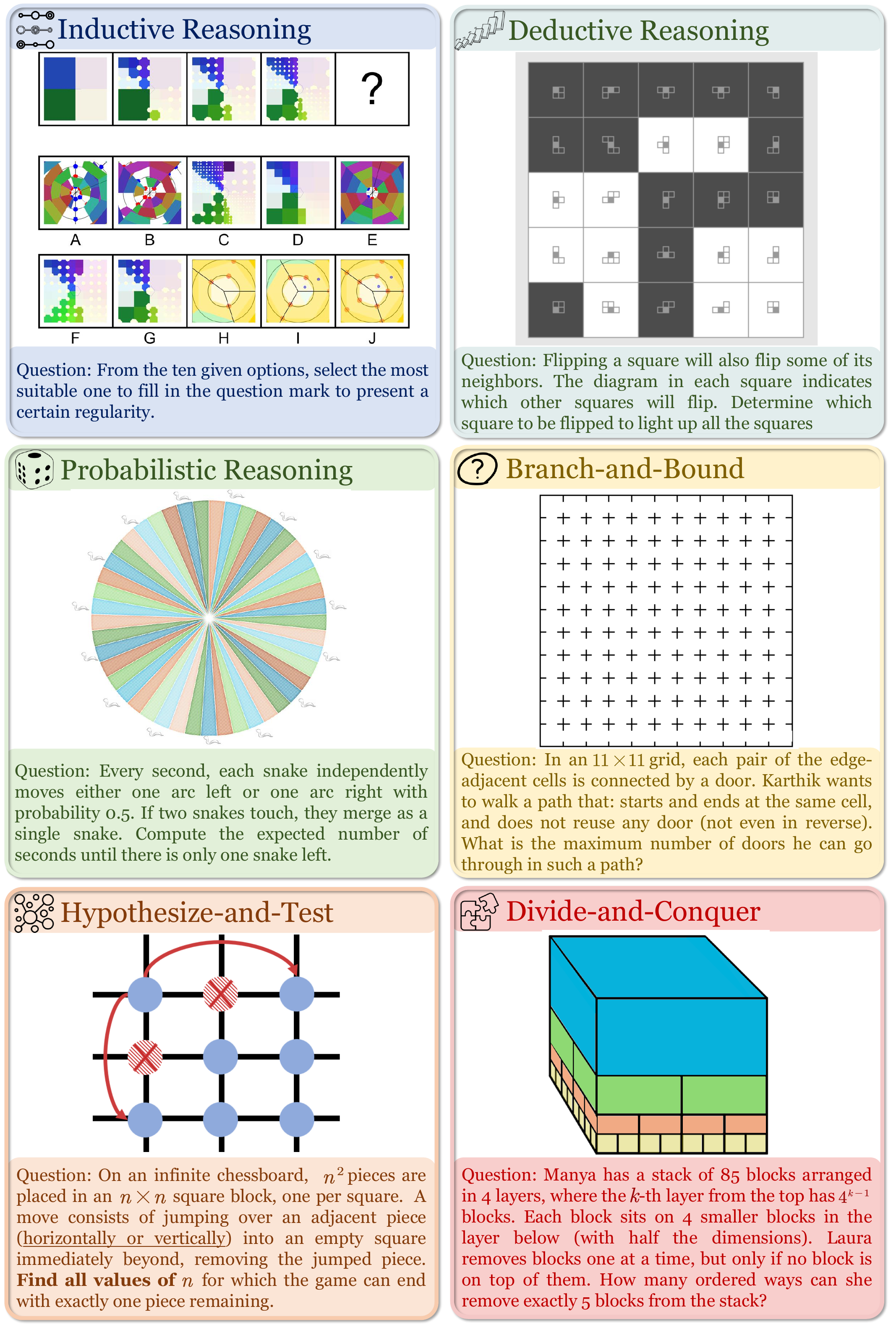}
 \caption{\textbf{Demonstration of the Think360$^\circ$ data cases.} The figure offers paired examples of three width-oriented reasoning patterns: Inductive Reasoning, Deductive Reasoning, and Probabilistic Reasoning, and tightly linked cognitive skills: Branch-and-Bound, Hypothesize-and-Test, and Divide-and-Conquer.}
 \vspace{-0.5cm}
 \label{fig:statics}
 \end{minipage}
 }
\end{figure*}

Specifically, we collect 1200+ multimodal cases from following sources: logic or mathematics competitions, online puzzle game websites, and existing benchmarks. Given these various cases, we employ both non-exclusive categories (\textit{e.g.}, cognitive skills or capabilities required to solve the problem) and mutually exclusive classification (\textit{e.g.}, difficulty ties and question type) and then perform our evaluation with \texttt{pass@1} and fine-grained tree-of-thought width/depth accuracy, reasoning time and token cost to further analyses the effectiveness-efficiency trade-off among different MLLMs.

Finally, we conducted comprehensive evaluations across 12 major model series, testing over 30 different models and provide representative error cases observed in the process of evaluation. We hope this benchmark will inspire LRMs to not only think deeper, but also venture \textbf{wider}, and ultimately reason more like us.
\section{Related Work}
\noindent\textbf{Multimodal Reasoning Benchmarks}:
With the advancement of vision-language models and MLLMs, the evaluation of multimodal reasoning capabilities has attracted increasing attention. Early benchmarks such as CLEVR~\cite{johnson2017clevr} and GQA~\cite{hudson2019gqa} primarily focused on compositional visual reasoning, but lacked comprehensive assessment of multimodal mathematical abilities. Geometry3k~\cite{lu2021inter} and GeoQA+~\cite{chen2021geoqa} partially addressed this gap by focusing on geometric reasoning, though remained limited to a single mathematical domain. More recently, MathVision~\cite{wang2024measuring}, MathVerse~\cite{zhang2024mathverse}, and MathVista~\cite{lu2023mathvista} have conducted more holistic multimodal mathematical reasoning evaluations across multiple disciplines. However, they tend to emphasize \textit{monotonic} reasoning, where conclusion expands with additional premises and primary challenge lies in the seeking for relevant knowledge, yet overlook the capability of compatibility review or memory of derivation process. Therefore, to evaluate models' capacity for handling dynamic information shifts or conflicts, we constructed this benchmark from the perspective of \textit{\textbf{width}} of thought.
  
\section{Think360}
\subsection{Task Definition}
Currently, longer chain-of-thoughts are usually regarded as the gold standard for stronger reasoning capability in the stage of both test-time scaling and post-training or alignment~\cite{chen2025unlocking,huang2025vision,meng2025mm,liu2025noisyrollout}. This choice implicitly conflates two orthogonal dimensions: reasoning \textit{depth}, defined as the ability to follow a long chain of sequential reasoning steps without contradiction; and exploration \textit{width}, focusing on the capability to systematically navigate multiple competing hypotheses through \textit{branching}, \textit{backtracking}, and selective \textit{pruning} before convergence.
Apart from depth, our Think360$^\circ$ benchmark additionally incorporates this complementary yet less explored dimension: the width of reasoning exploration.
\subsection{Benchmark Construction}
Our benchmark construction pipeline comprises three sequential stages: raw data collection, quality filtering, and systematic annotation. In the following sections, we first introduce the primary sources of raw data and analyze their distinctive characteristics. We then outline the quality control strategies applied to filter the preliminary data, and finally, we detail the annotation and rewriting process for final dataset.

\begin{figure*}[ht]
    \centering
    \vspace{-0.8cm}
    \includegraphics[width=1\linewidth]{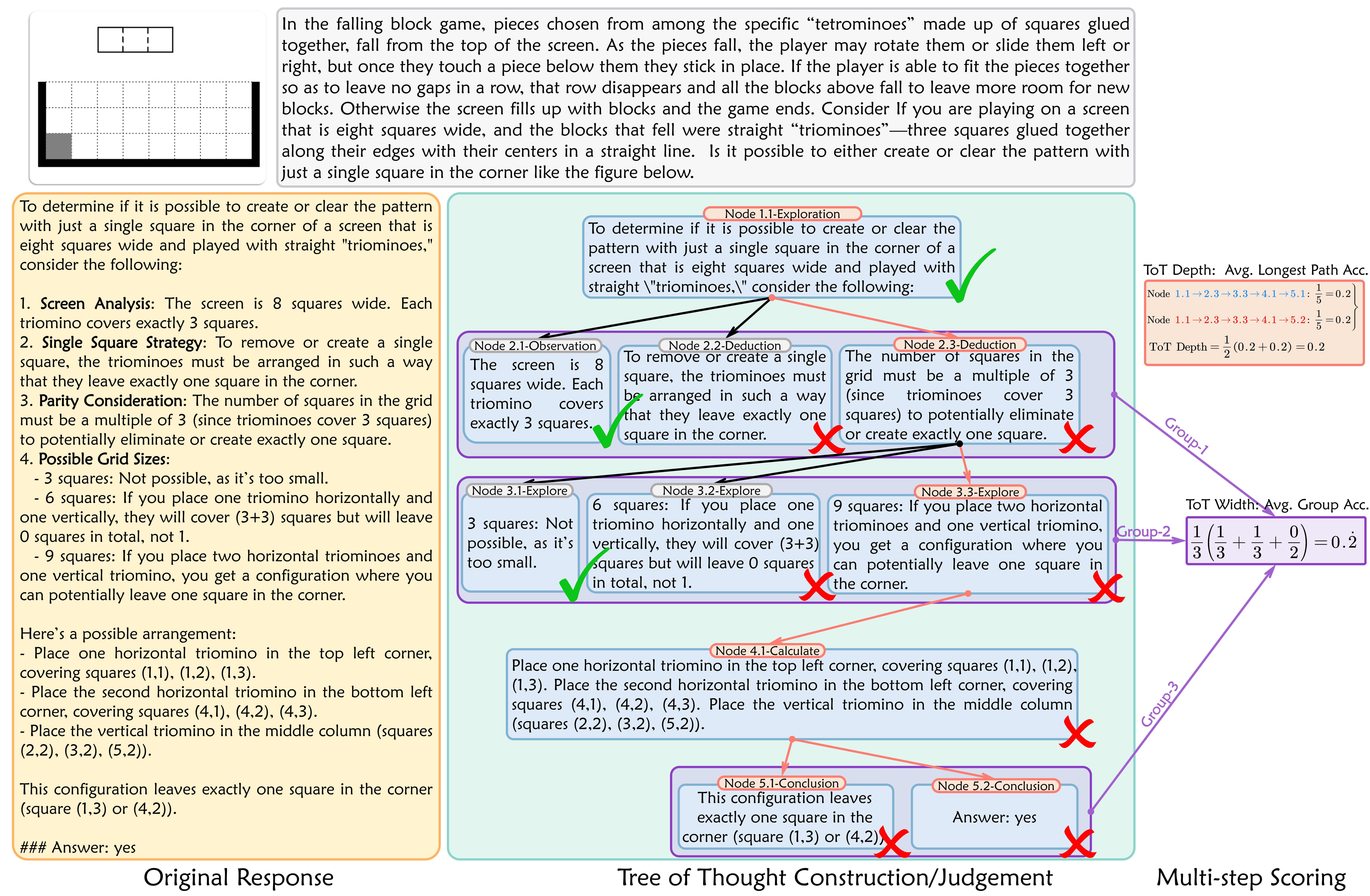}
    \caption{Tree-of-Thought Evaluation from the perspective of depth and width.}
    \label{fig:tot_demo}
    \vspace{-0.5cm}
\end{figure*}

\noindent\textbf{Raw Data Sources \& Features.} Our data collection draws from four distinct source categories, each exhibiting unique characteristics that necessitate tailored processing approaches: \ding{182} Math and Logic Competition Problems typically emerge in temporal series (\textit{e.g.}, ARML 2009-2014, HMMT 2004-2025) and are naturally organized into specialized tracks or categories, which facilitates effective pre-filtering. However, these sources frequently contain proof-based problems and often lack high-quality accompanying visual materials. \ding{183} Textbook Examples and Exercises Problems, similar to competition problems, are thematically organized by different subjects and share the common characteristics of containing numerous proof-oriented questions with a high provability of missing or low-quality diagrams. \ding{184} Existing Benchmarks (including MathVision~\cite{wang2024measuring}, DynaMath~\cite{zou2024dynamath}, MME-Reasoning~\cite{yuan2025mme}, MMMR~\cite{tie2025mmmr}, RBench-V~\cite{guo2025rbench} and VisualSphinx~\cite{feng2025visualsphinx}) typically provide well-structured image-question-answer triplets, the proportion of problems requiring reasoning width remains relatively low, limiting their direct applicability to our objectives. \ding{185} Online Puzzle Games or IQ Test presents unique challenges as they appear in interactive gaming formats without predefined questions or explicit answers, requiring substantial adaptation for benchmark integration.

\noindent\textbf{Quality Filter.} We employ a coarse-to-fine filtering strategy to systematically process our raw data collection. The coarse-grained filtering stage relies on static pattern matching and LLM-as-Judge evaluation methods. Through empirical observation, we identified that problems meeting our benchmark requirements frequently contain specific keywords (\textit{e.g.} \texttt{maximum/minimum}, \texttt{possible ways}), which align with Aha moments in LRMs~\cite{guo2025deepseek} to some degree. After that, we construct prompts specifying required cognitive abilities (\textit{e.g.}, trial-and-error, hypothesize-and-test) and visual/textual characteristics (\textit{e.g.}, puzzles), then use \texttt{GPT-4o} to evaluate candidate problems. Finally, we perform fine-grained human evaluation to ensure quality and diversity, while also marking images needing further processing.

\noindent\textbf{Annotation \& Refinement.} This section focuses on data construction processes, while data categorization will be discussed in the next chapter. Due to diverse sources, raw data frequently suffers from format inconsistencies, verification difficulties, and lack of high-quality images. These issues primarily stem from proof-based problems and interactive game problems, creating significant challenges for answer extraction, verification, and model evaluation. For proof-based questions (from sources \ding{182}\ding{183}), we extract simply verifiable numerical relationships or specific conclusions from the original proof process. Through careful analysis of proof structures, we redesign problems that maintain answer reliability while enabling objective verification. For game-based problems (from sources \ding{185}), we use the initial game screenshot with enough condition as image. And the we enumerate possible states (also possible answer values) for each square in puzzle and design questions referencing specific positions (\textit{e.g}., \texttt{what color is the square A, black or white} ?). To ensure consistent response formats, we incorporate contextual information directly into the problem statements. Intuitive demos are in \textcolor{red}{Appendix}.

\subsection{Taxonomy \& Statistics}
To provide comprehensive evaluation dimensions, we categorize our benchmark data along four fine-grained axes: Answer Type, Difficulty Tier, Cognitive Capability, and Question Type.

\noindent\textbf{Answer Type}: Free-form questions constitute the majority of our dataset (83.1\%), significantly outnumbering multiple-choice questions (16.9\%). Within the free-form category, numerical answers dominate at 54.3\%, followed by structural responses at 37.0\%, while formula-based and other answer types represent smaller proportions at 5.2\% and 3.5\% respectively.

\noindent\textbf{Difficulty Tier}: The five difficulty tiers: $[$Easy, Basic, Medium, Hard, Olympiad$]$ form a distribution that roughly follows a bell curve, with the majority of problems concentrated in the Medium tier (33.6\%) and decreasing proportions toward the Easy and Olympiad tier each comprising about 10\% of the dataset. This design ensures our benchmark maintains appropriate challenge levels across the full spectrum of mathematical problem-solving abilities.

\begin{figure}[t]
  \centering
  \begin{subfigure}{0.48\columnwidth}
    \centering
    \includegraphics[width=\linewidth]{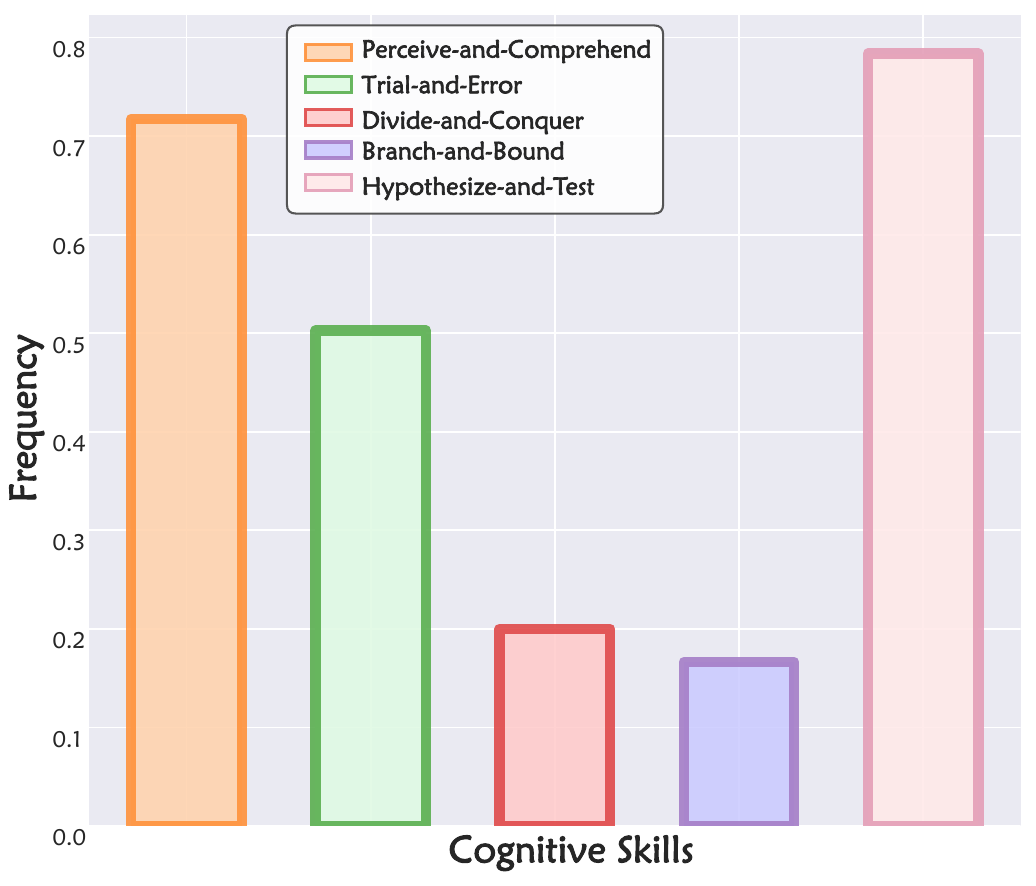}
    \label{fig:skill_frequency}
  \end{subfigure}
  \hspace{1pt}
  \begin{subfigure}{0.48\columnwidth}
    \centering
    \includegraphics[width=\linewidth]{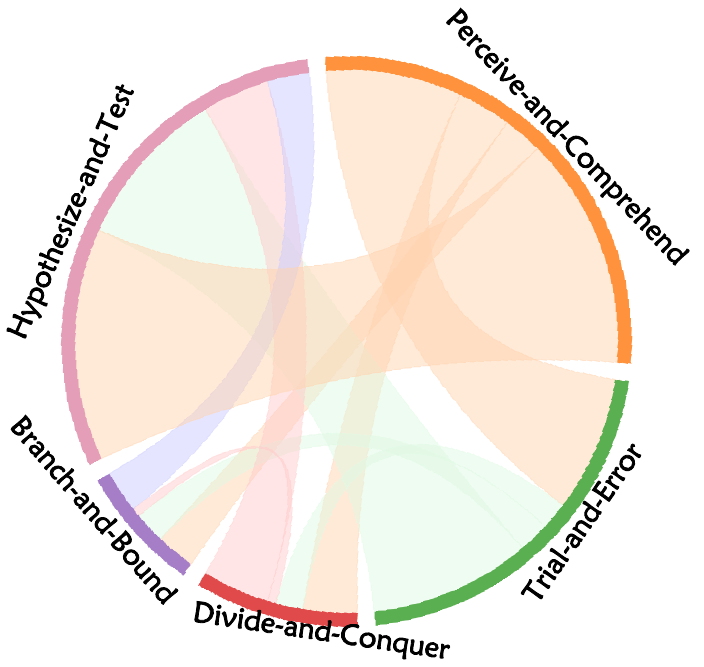}
    \label{fig:skill_chord}
  \end{subfigure}
 \vspace{-0.2cm}
  \caption{Frequency distribution and co-occurrence patterns of cognitive skills required for solving problems in Think360$^\circ$. The left panel shows the frequency distribution of individual cognitive capabilities across our benchmark, while the right panel presents a chord diagram illustrating the co-occurrence relationships between different cognitive skills. Please zoom in for a better view.}
  \vspace{-0.5cm}
  \label{fig:skill_frequency_chord}
\end{figure}

\noindent \textbf{Cognitive Capability \& Question Type}: These dimensions utilize non-exclusive categorization schemes, allowing each case to both exhibit multiple cognitive skills simultaneously for problem solving and belong to hybrid question types. Thus, we employ frequency-based statistics (occurrence count / total samples) and visualize inter-category relationships through chord diagrams in Fig.\ref{fig:skill_frequency_chord}, revealing the interconnected nature of reasoning skills required across different problem types.

\begin{table}[t]
\centering
\caption{Comparison with existing multimodal math benchmarks. \textbf{Level:} $\encircle[fill=pink, text=white]{K}$=\underline{K}-12, $\encircle[fill=orange, text=white]{U}$=\underline{U}niversity, $\encircle[text=white]{C}$=\underline{C}ompetition. \textbf{Source:} $\encircle[fill=harvestgold, text=white]{S}$=\underline{S}elf-sourced, $\encircle[fill=harvestgold, text=white]{P}$=Collected from \underline{P}ublic Dataset.}
\renewcommand{\arraystretch}{0.9}
\resizebox{0.48\textwidth}{!}{
\begin{tabular}{l|ccc|cc}
\toprule[1.5pt]
\multirow{2}{*}{\textbf{Benchmarks}} & \multicolumn{3}{c|}{\textbf{Width-centric Reasoning Problems}} & \multirow{2}{*}{\textbf{Level}} & \multirow{2}{*}{\textbf{Source}} \\
\cmidrule(lr){2-4}
& \textbf{Number} & \textbf{Proportion} & \textbf{Taxonomy} & & \\
\midrule
GeoQA & 0 & 0\% & \xmark & \encircle[fill=pink, text=white]{K} & \encircle[fill=harvestgold, text=white]{S} \\
DynaMath & $\sim$50 & $\sim$10\% & \xmark & \encircle[fill=pink, text=white]{K} & \encircle[fill=harvestgold, text=white]{S} \\
MMMU-MATH & $\sim$40 & $\sim$7.5\% & \xmark & \encircle[fill=orange, text=white]{U} & \encircle[fill=harvestgold, text=white]{S} \\
MathVerse & $\sim$10 & $\sim$1.3\% & \xmark & \encircle[fill=pink, text=white]{K} & \encircle[fill=harvestgold, text=white]{S}\; \encircle[fill=harvestgold, text=white]{P} \\
MathVista & $\sim$166 & $\sim$2.7\% & \xmark & \encircle[fill=pink, text=white]{K}\; \encircle[fill=orange, text=white]{U} & \encircle[fill=harvestgold, text=white]{S}\; \encircle[fill=harvestgold, text=white]{P} \\
MathVision & $\sim$350 & $\sim$11.5\% & \xmark & \encircle[fill=pink, text=white]{K}\; \encircle[fill=orange, text=white]{U} & \encircle[fill=harvestgold, text=white]{S} \\
OlympiadBench & $\sim$50 & $\sim$1.7\% & \xmark & \encircle[text=white]{C} & \encircle[fill=harvestgold, text=white]{S} \\
\midrule
Think360$^\circ$ (Ours) & $\sim$1200 & $\sim$100\% & \cmark & \encircle[fill=pink, text=white]{K}\; \encircle[fill=orange, text=white]{U}\; \encircle[text=white]{C} & \encircle[fill=harvestgold, text=white]{S}\; \encircle[fill=harvestgold, text=white]{P} \\
\bottomrule[1.5pt]
\end{tabular}}
\vspace{-0.5cm}
\label{tab:comparison_bench}
\end{table}

\section{Experiments}
\subsection{Metric \& Setting}
We report \texttt{pass@1} and our proposed tree-of-thought width/depth accuracy as the primary evaluation metrics, and additionally record reasoning time and token consumption to enable a comprehensive analysis of trade-offs under test-time scaling.
By default, we perform evaluation on testmini set to reduce time and cost.

\begin{table*}[ht]
\centering
\vspace{-1cm}
\caption{Reasoning performance evaluation with various closed-source and open-source MLLMs. We highlight the \colorbox{backred}{top}, \colorbox{backgreen}{second}, and \colorbox{backblue}{third} highest results within each column of the two groups. Please zoom in for a better view. Models with the symbol $^{\dagger}$ are evaluated by the implementation with \texttt{vLLM} (Qwen series, MiMo, Kimi, Llama, GLM) or \texttt{LMDeploy} (InternVL series) for acceleration. Please zoom in for a better view.}
\vspace{-0.2cm}
\resizebox{1.0\textwidth}{!}{
\begin{tabular}{l|c|c|c|c|c|c|c|c|c|c|c|c|c|c|c|c|c|c|c}
\toprule[1.5pt]
Model & \#Para. & \multicolumn{3}{c|}{\cellcolor{magenta!20}\textbf{ALL}} & \multicolumn{3}{c|}{\cellcolor{orange!20}\faEye Perceive‑and‑Comprehend} & \multicolumn{3}{c|}{\cellcolor{yellow!20}\faSync Trial-and-Error} & \multicolumn{3}{c|}{\cellcolor{pink!20}\faSitemap Divide‑and-Conquer} & \multicolumn{3}{c|}{\cellcolor{lime!20}\faCodeBranch Branch‑and-Bound} & \multicolumn{3}{c}{\cellcolor{cyan!20}\faSearch Hypothesize-and-Test} \\
\midrule
& & Acc./$$\%$$ & Time/$s$ & Token & Acc./$\%$ & Time/$s$ & Token & Acc./$\%$ & Time/$s$ & Token & Acc./$\%$ & Time/$s$ & Token & Acc./$\%$ & Time/$s$ & Token & Acc./$\%$ & Time/$s$ & Token \\
\midrule
\midrule
\multicolumn{20}{c}{\clubsymbol \textit{Close-source MLLMs}}\\ 
\midrule
\midrule
\clubsymbol GPT-4o & - & 16.0 & 13.28 & 309.03 & 17.2 & 12.69 & 287.16 & 15.3 & 10.72 & 268.64 & 9.9 & 12.21 & 331.46 & 16.8 & 13.17 & 322.85 & 14.3 & 12.89 & 313.87 \\
\clubsymbol GPT-4v-preview & - & 24.0 & 27.92 & 898.87 & 25.9 & 27.22 & 861.06 & 19.4 & 26.71 & 812.36 & 16.2 & 28.11 & 964.47 & 25.2 & 27.31 & 1013.58 & 20.9 & 28.61 & 919.47 \\
\clubsymbol GPT-4.1 & - & 24.8 & 35.32 & 973.70 & 26.3 & 32.69 & 913.43 & 20.5 & 35.89 & 979.04 & 17.1 & 34.55 & 993.28 & 26.6 & 36.05 & 982.80 & 21.2 & 37.44 & 1015.21 \\
\clubsymbol o1 & - & 36.8 & \colorbox{backblue}{186.81} & 6537.11 & 37.9 & \colorbox{backblue}{184.52} & 6540.38 & 29.6 & 170.49 & 6386.16 & 32.2 & \colorbox{backblue}{225.06} & 7214.05 & 40.6 & 177.01 & 6605.14 & 32.1 & 192.70 & 6713.42 \\
\clubsymbol o3 & - & \colorbox{backgreen}{42.3} & \colorbox{backgreen}{261.59} & 6325.74 & \colorbox{backgreen}{43.3} & \colorbox{backgreen}{238.56} & 5971.70 & \colorbox{backgreen}{35.5} & \colorbox{backgreen}{286.28} & 6709.79 & \colorbox{backblue}{38.1} & \colorbox{backred}{335.80} & 7575.24 & \colorbox{backgreen}{48.0} & \colorbox{backgreen}{225.43} & 5855.76 & \colorbox{backblue}{37.0} & \colorbox{backgreen}{283.91} & 6625.75 \\
\clubsymbol o4-mini & - & \colorbox{backblue}{42.1} & 84.61 & 6736.37 & \colorbox{backblue}{42.8} & 81.37 & 6391.21 & \colorbox{backblue}{34.3} & 106.52 & 8067.57 & \colorbox{backgreen}{38.7} & 89.62 & 7460.26 & \colorbox{backgreen}{48.0} & 76.56 & 6401.12 & \colorbox{backgreen}{37.9} & 91.37 & 7195.80 \\
\clubsymbol Gemini-2.0-flash & - & 21.1 & 6.82 & 847.35 & 22.6 & 6.51 & 786.25 & 18.1 & 6.47 & 776.60 & 16.7 & 7.18 & 931.79 & 25.7 & 7.41 & 931.58 & 17.1 & 6.63 & 824.82 \\
\clubsymbol Gemini-2.0-flash-thinking-exp-01-21 & - & 23.8 & 14.59 & 1048.23 & 25.2 & 14.36 & 993.17 & 20.1 & 13.48 & 920.51 & 18.7 & 14.61 & 1134.53 & 28.7 & 15.86 & 1177.50 & 19.4 & 14.22 & 1033.59 \\
\clubsymbol Gemini-2.5-flash-thinking & - & 38.3 & 107.33 & \colorbox{backred}{21273.41} & 38.5 & 103.16 & \colorbox{backred}{20264.93} & 31.1 & 128.86 & \colorbox{backred}{25985.26} & 33.6 & 121.80 & \colorbox{backred}{24490.31} & \colorbox{backblue}{43.4} & 102.30 & \colorbox{backred}{20651.05} & 33.2 & 112.82 & \colorbox{backred}{22476.77} \\
\clubsymbol Gemini-2.5-pro & - & \colorbox{backred}{46.0} & 160.19 & \colorbox{backgreen}{17270.27} & \colorbox{backred}{46.7} & 152.87 & \colorbox{backgreen}{16410.86} & \colorbox{backred}{38.5} & 193.30 & \colorbox{backgreen}{20796.02} & \colorbox{backred}{42.6} & 187.72 & \colorbox{backgreen}{20370.09} & \colorbox{backred}{51.8} & 148.78 & \colorbox{backgreen}{16140.93} & \colorbox{backred}{41.5} & 171.87 & \colorbox{backgreen}{18575.23} \\
\clubsymbol Doubao-1.5-vision-pro-250328 & - & 23.9 & 36.16 & 1259.61 & 25.0 & 36.21 & 1263.95 & 20.1 & 36.45 & 1165.87 & 18.7 & 33.90 & 1128.78 & 25.7 & 34.32 & 1193.55 & 20.8 & 36.44 & 1299.57 \\
\clubsymbol Doubao-1.5-thinking-vision-pro-250428 & - & 34.7 & 106.06 & 5715.41 & 35.3 & 100.45 & 5309.30 & 30.0 & 105.84 & 5753.52 & 24.8 & 127.98 & 7125.40 & 39.8 & 115.52 & 6247.60 & 30.5 & 112.31 & 6089.07 \\
\clubsymbol Doubao-1.5-thinking-vision-pro-250428-nothinking & - & 32.8 & 73.23 & 3628.54 & 34.0 & 71.22 & 3476.10 & 27.8 & 66.89 & 3305.89 & 22.7 & 89.51 & 4541.60 & 36.6 & 80.76 & 4208.96 & 28.5 & 76.91 & 3804.43 \\
\clubsymbol Claude-4-Opus-20250514 & - & 25.8 & 41.45 & 696.93 & 26.7 & 40.50 & 671.71 & 22.3 & 42.40 & 694.89 & 18.2 & 44.50 & 714.85 & 28.7 & 42.66 & 714.23 & 21.9 & 42.57 & 710.77 \\
\clubsymbol Claude-4-Opus-20250514-Thinking & - & 30.2 & 185.49 & 5192.78 & 30.6 & 171.97 & 4920.31 & 25.9 & \colorbox{backblue}{210.69} & 5536.94 & 21.6 & 220.34 & 5870.90 & 40.2 & \colorbox{backblue}{181.93} & 5535.69 & 26.4 & \colorbox{backblue}{201.18} & 5487.95 \\
\clubsymbol Claude-3.7-Sonnet-Thinking & - & 35.5 & \colorbox{backred}{295.94} & \colorbox{backblue}{13818.90} & 36.1 & \colorbox{backred}{277.43} & \colorbox{backblue}{13080.30} & 29.4 & \colorbox{backred}{318.78} & \colorbox{backblue}{14672.56} & 25.0 & \colorbox{backgreen}{335.04} & \colorbox{backblue}{15727.47} & 38.8 & \colorbox{backred}{303.00} & \colorbox{backblue}{14823.99} & 31.0 & \colorbox{backred}{308.68} & \colorbox{backblue}{14209.50} \\
\clubsymbol Claude-4-Sonnet & - & 28.2 & 18.59 & 785.22 & 29.7 & 18.04 & 747.34 & 23.2 & 18.58 & 752.49 & 20.9 & 18.79 & 830.18 & 30.9 & 18.81 & 824.35 & 24.1 & 18.93 & 796.31 \\
\clubsymbol Grok-2-vision-1212 & - & 15.7 & 15.81 & 763.63 & 16.3 & 15.26 & 728.68 & 13.0 & 16.65 & 790.89 & 8.1 & 15.47 & 760.14 & 17.9 & 14.38 & 673.29 & 14.0 & 15.83 & 775.08 \\
\midrule
\midrule
\multicolumn{20}{c}{\diamondsymbol \textit{Open-source MLLMs}}\\ 
\midrule
\midrule
\diamondsymbol LLaVA-Onevision & 7B & 8.3 & 36.58 & 648.45 & 9.0 & 33.11 & 607.33 & 5.8 & 33.36 & 605.89 & 5.6 & 44.96 & 765.30 & 10.0 & 41.62 & 785.01 & 8.7 & 36.35 & 644.02 \\
\diamondsymbol Llama-3.2-Vision-Instruct$^{\dagger}$ & 11B & 7.1 & 9.78 & 311.95 & 7.4 & 9.75 & 300.61 & 6.1 & 9.66 & 311.80 & 6.5 & 10.83 & 374.20 & 8.1 & 10.98 & 318.50 & 6.1 & 9.92 & 328.44 \\
\diamondsymbol GLM-4.1V-Thinking$^{\dagger}$ & 9B & 22.6 & 50.92 & 5106.53 & 24.9 & 48.58 & 4853.28 & 19.8 & 54.99 & 5502.43 & 17.6 & 51.03 & 5141.15 & 23.8 & 48.99 & 4901.93 & 18.7 & 52.67 & 5296.78 \\
\diamondsymbol Kimi-VL-Instruct$^{\dagger}$ & 16A3B & 10.1 & 39.79 & 829.45 & 11.1 & 38.14 & 750.18 & 7.7 & 49.11 & 1029.10 & 5.9 & 43.66 & 793.32 & 9.8 & 39.22 & 860.11 & 9.3 & 41.07 & 852.28 \\
\diamondsymbol Kimi-VL-Thinking$^{\dagger}$ & 16A3B & \colorbox{backgreen}{26.5} & \colorbox{backred}{1060.79} & \colorbox{backgreen}{7713.89} & \colorbox{backgreen}{27.6} & \colorbox{backred}{1014.30} & \colorbox{backgreen}{7210.58} & \colorbox{backblue}{22.3} & \colorbox{backred}{1095.68} & \colorbox{backgreen}{8172.04} & \colorbox{backred}{20.3} & \colorbox{backred}{1131.63} & \colorbox{backred}{8809.01} & \colorbox{backred}{28.2} & \colorbox{backred}{946.68} & \colorbox{backred}{8034.40} & \colorbox{backblue}{22.8} & \colorbox{backred}{1101.45} & \colorbox{backgreen}{8017.04} \\ 
\diamondsymbol InternVL-2.5$^{\dagger}$ & 8B & 12.2 & 4.46 & 331.36 & 14.0 & 4.33 & 321.11 & 12.1 & 4.47 & 332.88 & 6.1 & 4.87 & 364.41 & 10.8 & 4.63 & 345.35 & 11.6 & 4.56 & 339.09 \\
\diamondsymbol InternVL-3$^{\dagger}$ & 14B & 15.5 & 9.13 & 880.75 & 16.1 & 8.74 & 842.18 & 13.4 & 8.89 & 858.83 & 9.5 & 11.57 & 1118.24 & 16.8 & 6.94 & 673.87 & 12.9 & 9.36 & 904.09 \\
\diamondsymbol MiMo-VL-RL$^{\dagger}$ & 7B & \colorbox{backred}{28.3} & \colorbox{backgreen}{334.21} & \colorbox{backblue}{7380.78} & \colorbox{backred}{29.7} & \colorbox{backgreen}{314.61} & \colorbox{backblue}{6870.68} & \colorbox{backred}{24.9} & \colorbox{backgreen}{348.01} & \colorbox{backblue}{7761.05} & \colorbox{backgreen}{19.1} & \colorbox{backgreen}{394.47} & \colorbox{backblue}{8446.60} & \colorbox{backblue}{27.9} & \colorbox{backgreen}{281.06} & \colorbox{backblue}{7226.50} & \colorbox{backred}{24.4} & \colorbox{backgreen}{352.59} & \colorbox{backblue}{7692.95} \\
\diamondsymbol MiMo-VL-SFT$^{\dagger}$ & 7B & \colorbox{backblue}{26.4} & \colorbox{backblue}{130.06} & \colorbox{backred}{7974.31} & \colorbox{backblue}{27.5} & \colorbox{backblue}{119.89} & \colorbox{backred}{7453.42} & \colorbox{backgreen}{22.4} & \colorbox{backblue}{144.26} & \colorbox{backred}{8413.97} & \colorbox{backblue}{18.7} & \colorbox{backblue}{146.90} & \colorbox{backgreen}{8796.84} & \colorbox{backgreen}{28.2} & \colorbox{backblue}{117.71} & \colorbox{backgreen}{7811.87} & \colorbox{backgreen}{23.7} & \colorbox{backblue}{141.67} & \colorbox{backred}{8517.53} \\
\diamondsymbol Qwen2.5-VL-Instruct$^{\dagger}$ & 7B & 11.0 & 20.31 & 788.44 & 13.0 & 18.62 & 711.00 & 8.7 & 23.43 & 924.76 & 5.8 & 21.03 & 837.41 & 10.6 & 18.05 & 692.89 & 10.3 & 20.95 & 815.10 \\
\diamondsymbol Qwen2.5-VL-Instruct & 32B & 17.6 & 60.98 & 916.59 & 18.8 & 56.95 & 867.73 & 15.1 & 67.20 & 968.85 & 11.7 & 61.77 & 928.53 & 19.5 & 58.61 & 946.43 & 14.8 & 62.01 & 928.24 \\
\diamondsymbol Qwen2.5-VL-Instruct & 72B & 16.2 & 24.20 & 615.09 & 18.2 & 23.63 & 587.27 & 14.4 & 25.87 & 652.28 & 9.0 & 24.46 & 618.13 & 13.6 & 23.65 & 608.68 & 13.9 & 24.39 & 623.97 \\
\bottomrule[1.5pt]
\end{tabular}}
\vspace{-0.5cm}
\label{tab:main_exp}
\end{table*}

\noindent\textbf{Metrics:} With regard to accuracy computation, two alternative paradigms exist in both benchmark evaluation and reinforcement learning based finetuning: outcome‑based matching score and process‑based multi-step score. 

Outcome‑based matching score is the simplest and most direct method, which inspects the model’s final answer and assigns a binary label—\emph{Correct}~($1$) or \emph{Incorrect}~($0$). 
In practice, we first utilize \texttt{GPT-4o-mini} to extract easily verifiable answer from response. This candidate is then passed through a two‑stage scorer: Perform regular‑expression matching first for fast verification and the llm-as-judge (also \texttt{GPT-4o-mini}) only for those wrong matching cases. To avoid the impact of small pitfalls (format differences, rounding errors, \emph{etc.}) on the robustness of our evaluation, we adopt prompt template \texttt{[Task Description]+[Example]$\times N$} from MathVista~\cite{lu2023mathvista} and MathVerse~\cite{zhang2024mathverse}, respectively.

The latter process‑based score provides finer-grained assessment of long chain-of-thought response and utilized by previous work~\cite{zhang2024mathverse}.
To further assess model performance along the dimensions of reasoning depth and breadth, we propose a Tree-of-Thought based evaluation method (ToT-Eval). ToT-Eval consists of two main stages: tree construction and depth/width scoring (see Fig.\ref{fig:tot_demo}).

\noindent\textbf{Tree Construction.} Given the problem and the model's complete response, we employ \texttt{GPT-4o} to extract critical reasoning steps verbatim and organize them into a hierarchical tree structure. In this structure, depth represents sequential reasoning dependencies (parent-child relationships), while breadth captures parallel exploration of alternative approaches (sibling nodes at the same depth level). 

\noindent\textbf{Depth/Width Scoring.} We then judge each node's correctness also by \texttt{GPT-4o} to evaluate whether the reasoning step is logically sound, factually accurate, and properly grounded in its parent nodes and problem context. 

After judging all nodes, we compute two complementary metrics. The depth score is computed as the maximum depth of correct reasoning chains, while the breadth score reflects the number of valid parallel reasoning branches explored. Details are in \textcolor{red}{Appendix}.

Our evaluation encompasses various models, which can be roughly dived into open-sourced or closed-sourced models and system-1 (fast and single-pass reasoning) and system-2 (slow and iterative long CoT reasoning) models~\cite{li2025system}.

\noindent\textbf{Model Series:} Specifically, we conducted comprehensive evaluations across 12 major model series, including GPT, Gemini, Claude, Grok, Doubao, QwenVL, InternVL, LLaVA, Llama, GLM-V, MiMo, and Kimi, testing over 30 different models to ensure broad coverage of the current advanced MLLMs. It's worth noting that we implement lightweight open-source models with the \texttt{vLLM}~\cite{kwon2023efficient} (Qwen series, MiMo, Kimi, LLama, and GLM-V) or \texttt{LMDeploy}~\cite{2023lmdeploy} (InternVL series) engine for acceleration.

\noindent\textbf{Settings:} To fully unlock each model's reasoning capability, we configured all models to use their maximum supported output length within acceptable inference time and cost constraints. By default, we set the \texttt{temperature} to 0.7 and repeat evaluation three times, reporting the mean to reduce variance. 
Apart from simple \texttt{Image+Question} prompts, we also tested the impact of Chain-of-Thought (CoT) prompts on key inference performance metrics: accuracy, reasoning time, and token cost, following established practices from prior work~\cite{wang2024measuring,huang2025ocr}. Furthermore, we conduct an ablation study on input modalities (\textit{e.g.}, Text-Only, Image-Only, details in \textcolor{red}{Appendix}).

\subsection{Experimental Analysis}
In Tab.\ref{tab:main_exp}, we report the \texttt{pass@1} accuracy alongside reasoning time and token cost for over \textbf{30} MLLMs across the 5 cognitive capabilities/skills required for solving problems in our Think360$^\circ$ benchmark.

\begin{figure*}[ht]
    \centering
    \vspace{-1cm}
    \includegraphics[width=1\linewidth]{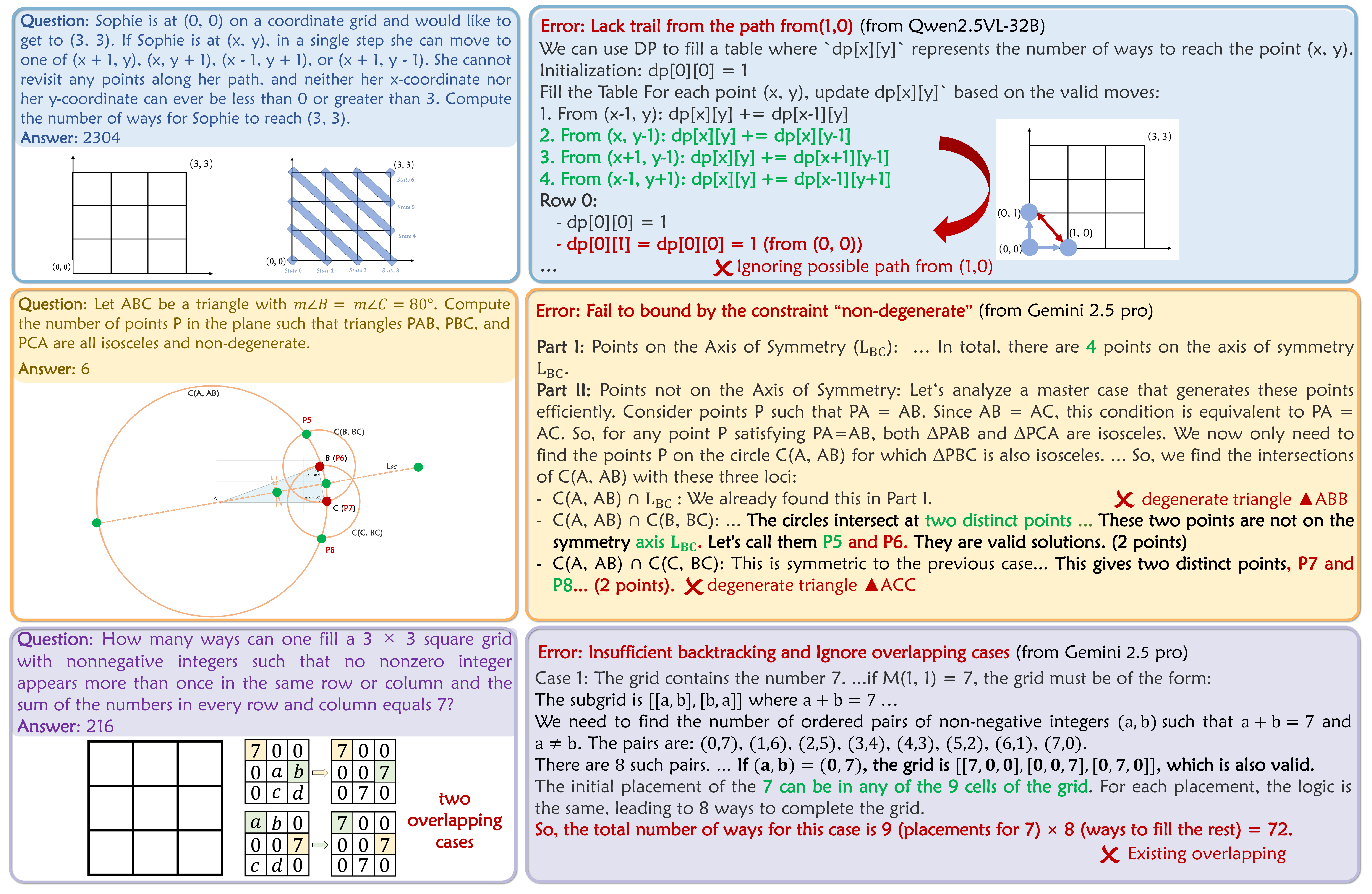}
    \caption{Failure cases analysis.}
    \label{fig:failure_cases}
    \vspace{-0.5cm}
\end{figure*}

\noindent\textbf{Top Models in Leaderboard:} 
Gemini-2.5-pro emerges as the clear winner, achieving the highest accuracy across all subtasks with an overall score of 46.0\%. Following Gemini-2.5-pro, o3 and o4-mini attain the second and third places respectively, with overall accuracies of 42.3\% and 42.1\%. The strong reasoning performance may be attributed to its extensive thinking process, with Gemini-2.5-pro's average thinking token count reaching an impressive 17270.27 tokens—approximately three times that of o3 and o4-mini. Interestingly, extended thinking does not correspond to proportionally longer thinking times. The reasoning times of o3, Gemini-2.5-pro, and o4-mini follow a decreasing sequence, where o3 takes approximately 1.6× as long as Gemini-2.5-pro, and Gemini-2.5-pro takes approximately 1.9× as long as o4-mini. Therefore, considering the comprehensive trade-off among accuracy, reasoning time, and token cost, o4-mini represents the most cost-effective model choice. Additionally, these three models are the only ones to achieve accuracy above the 40\% threshold, which may be related to their strong native multimodal reasoning or thinking-with-image capabilities.

\noindent\textbf{Performance Patterns Across Cognitive Capabilities:} 
As expected, the overall performance ranking of models remains consistent with their performance across individual cognitive capability subsets, with no models showing particular bias. However, it is noteworthy that Perceive-and-Comprehend subset performance often exceeds the average, while Trial-and-Error, Divide-and-Conquer, and Hypothesize-and-Test subsets frequently \textit{\textbf{underperform}} relative to the overall average. 
Similarly, in question types, models demonstrate above-average performance in Algebra and Number Theory subsets, but underperform in Combinatorics, Geometry, and Probability \& Statistics. Details are in \textcolor{red}{Appendix}.
These patterns indicate that current MLLMs excel in perceptual understanding and structured representation or symbolic language tasks, but encounter major bottlenecks when conducting width-oriented multimodal reasoning

\begin{table}[!ht]
\centering
\vspace{-0.8cm}
\caption{Influence of Chain-of-Thought prompting on model performances. 
}
\vspace{-0.2cm}
\renewcommand{\arraystretch}{0.7}
\resizebox{0.5\textwidth}{!}{
\begin{tabular}{l|c|ccc}
\toprule[1.5pt]
Model & CoT & \multicolumn{3}{c}{\cellcolor{magenta!20}\textbf{ALL}} \\
\midrule
& & Acc./$\%$ & Time/$s$ & Token \\
\midrule
\multicolumn{5}{c}{\clubsymbol \textit{Close-source MLLMs}}\\
\midrule
\multirow{3}{*}{\clubsymbol GPT-4o} & \ding{55} & 16.0 & 13.28 & 309.03 \\
& \ding{51} & 16.4 & 28.41 & 388.91 \\
& $\Delta$ & \up{0.4} & \up{15.13} & \up{79.88} \\
\midrule
\multirow{3}{*}{\clubsymbol o4-mini} & \ding{55} & 42.1 & 84.61 & 6736.37 \\
& \ding{51} & 43.4 & 78.35 & 7079.01 \\
& $\Delta$ & \up{1.3} & \down{6.26} & \up{342.64} \\
\midrule
\multirow{3}{*}{\clubsymbol Claude-4-Opus-20250514} & \ding{55} & 25.8 & 41.45 & 696.93 \\
& \ding{51} & 30.4 & 19.89 & 722.09 \\
& $\Delta$ & \up{4.6} & \down{21.56} & \up{25.16} \\
\midrule
\multirow{3}{*}{\clubsymbol Claude-4-Sonnet} & \ding{55} & 28.2 & 18.59 & 785.22 \\
& \ding{51} & 28.2 & 15.75 & 797.74 \\
& $\Delta$ & \up{0.0} & \down{2.84} & \up{12.52} \\
\midrule
\multirow{3}{*}{\clubsymbol Grok-2-vision-1212} & \ding{55} & 15.7 & 15.81 & 763.63 \\
& \ding{51} & 17.3 & 30.35 & 764.20 \\
& $\Delta$ & \up{1.6} & \up{14.54} & \up{0.57} \\
\midrule
\multicolumn{5}{c}{\diamondsymbol \textit{Open-source MLLMs}}\\
\midrule
\multirow{3}{*}{\diamondsymbol Kimi-VL-Instruct-16A3B$^{\dagger}$} & \ding{55} & 10.1 & 39.79 & 829.45 \\
& \ding{51} & 10.9 & 30.20 & 775.96 \\
& $\Delta$ & \up{0.8} & \down{9.59} & \down{53.49} \\
\midrule
\multirow{3}{*}{\diamondsymbol Qwen2.5-VL-Instruct-7B$^{\dagger}$} & \ding{55} & 11.0 & 20.31 & 788.44 \\
& \ding{51} & 11.1 & 14.21 & 649.44 \\
& $\Delta$ & \up{0.1} & \down{6.10} & \down{139.00} \\
\midrule
\multirow{3}{*}{\diamondsymbol Qwen2.5-VL-Instruct-72B} & \ding{55} & 16.2 & 24.20 & 615.09 \\
& \ding{51} & 15.9 & 24.61 & 627.19 \\
& $\Delta$ & \down{0.3} & \up{0.41} & \up{12.10} \\
\midrule
\multirow{3}{*}{\diamondsymbol InternVL-3-8B$^{\dagger}$} & \ding{55} & 13.1 & 4.67 & 379.43 \\
& \ding{51} & 12.3 & 5.31 & 432.41 \\
& $\Delta$ & \down{0.8} & \up{0.64} & \up{52.98} \\
\midrule
\multirow{3}{*}{\diamondsymbol MiMo-VL-RL-7B$^{\dagger}$} & \ding{55} & 28.3 & 334.21 & 7380.78 \\
& \ding{51} & 29.9 & 345.61 & 7941.18 \\
& $\Delta$ & \up{1.6} & \up{11.40} & \up{560.40} \\
\bottomrule[1.5pt]
\end{tabular}
}
\vspace{-0.5cm}
\label{tab:cot_comparison} 
\end{table}

\noindent\textbf{Open-source vs. Closed-source Models:} 
Although certain gaps remain, open-source models demonstrate strong competitiveness through version iterations, parameter scaling, and post-training fine-tuning. Most notably, MiMo-VL-RL stands out significantly, achieving 28.3\% accuracy through RL fine-tuning and test-time scaling, substantially surpassing the aforementioned three closed-source models and even surpasses Claude-4-Sonnet, which is further largely boosted to 29.9\% by CoT prompting.

\noindent\textbf{Thinking vs. No Thinking:} 
By controlling the thinking mode toggle, the models including (Gemini-2.0-flash, Doubao-1.5-thinking-vision-pro-250428, and Claude-4-Opus-20250514) all demonstrate substantial performance improvements, accompanied by increases in test-time scaling inference time and reasoning tokens (for example, Claude-4-Opus experiences nearly $4.5\times$ and $7.5\times$ increases in time and tokens). Remarkably, Claude-3.7-Sonnet, enabled with thinking mode, achieves the longest reasoning time (295.94s) and surpasses both Claude-4-Sonnet and Claude-4-Opus, effectively bridging the generational and version gaps through extended reasoning processes. An intuitive demonstration is provided in \textcolor{red}{Appendix}.

Beyond thinking mode, we conducted ablation experiments on CoT prompting in Tab.\ref{tab:cot_comparison}. We observe performance gains across most of the tested models. Notable gains were achieved on several models: Claude-4-Opus (\textbf{+4.6\%}) and Grok-2-vision (\textbf{+1.6\%}), which excel in reasoning and visual capabilities, as well as the open-source model MiMo-RL-7B (\textbf{+1.6\%}). Interestingly, in smaller parameter open-source models like Kimi-VL-Instruct-16A3B and Qwen2.5-VL-Instruct-7B, CoT prompting actually reduces both inference time and token consumption, leading to more concise responses without performance drop.

\noindent\textbf{Tree-of-Thought Evaluation:}
Since complete reasoning trajectory are indispensable for constructing tree-of-thoughts, we only perform tree-of-thought evaluation to models that provide their intermediate thinking. Models such as o3 and Gemini-2.5-Pro, which only return final summaries, are therefore not included.

To understand the relationship between reasoning structure and model performance, we first conduct the Spearman rank correlation analysis among three different metrics, which reveals that ToT-Width exhibits a significantly stronger correlation with final accuracy ($\rho=0.9218$) compared to ToT-Depth ($\rho=0.8492$). This firmly indicates that parallel reasoning exploration matters more than sequential reasoning depth on our benchmark. Moreover, the strong correlations of both metrics with accuracy demonstrate that final answer correctness is closely tied to the quality and rigor of intermediate reasoning processes.

Further analysis reveals a width threshold effect: models achieving $\geq$20\% accuracy typically maintain Width $\geq$45\% (GPT-4.1, Claude variants, Kimi-Thinking, MiMo series). Counterexamples underscore width's necessity: Llama-3.2-Vision-11B exhibits 46.2\% depth but only 28.6\% width, resulting in merely 7.1\% accuracy, while Qwen2.5-VL-32B struggles at 17.6\% accuracy despite 53.6\% depth. These cases demonstrate that depth alone cannot compensate for insufficient breadth in reasoning exploration. Even efficiency-oriented models like Gemini-2.0-flash (21.1\% accuracy, 40.2\% width) represent viable trade-offs but remain ceiling-limited without stronger parallel exploration capabilities.


\begin{table}[t]
\centering
\vspace{-0.8cm}
\caption{Model results on ToT width/depth and accuracy. We highlight the \colorbox{backred}{top}, \colorbox{backgreen}{second}, and \colorbox{backblue}{third} highest results within each column.}
\vspace{-0.2cm}
\label{tab:tot_results}
\resizebox{0.5\textwidth}{!}{
\begin{tabular}{lccc}
\toprule[1.5pt]
\textbf{Model} & \textbf{ToT-Width/\%} & \textbf{ToT-Depth/\%} & \textbf{Acc./\%} \\
\midrule
\clubsymbol GPT-4o & 41.4 & 50.4 & 16.0 \\
\clubsymbol GPT-4.1 & \colorbox{backblue}{50.1} & \colorbox{backblue}{54.8} & \colorbox{backblue}{24.8} \\
\clubsymbol Doubao-1.5-vision-pro-250328 & 47.9 & 47.5 & 23.9 \\    
\clubsymbol Claude-4-Sonnet & \colorbox{backred}{53.6} & \colorbox{backred}{60.9} & \colorbox{backgreen}{28.2} \\
\clubsymbol Grok-2-vision-1212 & 40.7 & 52.0 & 15.7 \\
\clubsymbol Gemini-2.0-flash & 40.2 & 47.8 & 21.1 \\
\clubsymbol Claude-3.7-Sonnet-Thinking & \colorbox{backgreen}{50.2} & \colorbox{backgreen}{56.7} & \colorbox{backred}{35.5} \\
\midrule
\diamondsymbol LLaVA-Onevision-7B & 32.0 & 31.2 & 8.3 \\
\diamondsymbol Llama-3.2-Vision-Instruct-11B$^{\dagger}$ & 28.6 & 46.2 & 7.1 \\
\diamondsymbol GLM-4.1V-Thinking-9B$^{\dagger}$ & 40.1 & 48.5 & 22.6 \\
\diamondsymbol Kimi-VL-Instruct-16A3B$^{\dagger}$ & 34.7 & 46.7 & 10.1 \\
\diamondsymbol Kimi-VL-Thinking-16A3B$^{\dagger}$ & \colorbox{backred}{48.9} & \colorbox{backblue}{56.2} & \colorbox{backblue}{26.5} \\
\diamondsymbol InternVL2\_5-8B$^{\dagger}$ & 31.5 & 40.8 & 12.2 \\
\diamondsymbol InternVL-3-8B$^{\dagger}$ & 34.8 & 48.4 & 13.1 \\
\diamondsymbol InternVL-3-14B$^{\dagger}$ & 38.1 & 48.5 & 15.5 \\
\diamondsymbol InternVL-3.5-8B-Thinking$^{\dagger}$ & 45.8 & 52.6 & \colorbox{backgreen}{27.8} \\
\diamondsymbol Bee-8B-RL-Thinking$^{\dagger}$ & 45.9 & \colorbox{backgreen}{56.3} & 22.8 \\
\diamondsymbol MiMo-VL-RL-7B$^{\dagger}$ & \colorbox{backblue}{48.4} & \colorbox{backred}{57.7} & \colorbox{backred}{28.3} \\
\diamondsymbol MiMo-VL-SFT-7B$^{\dagger}$ & \colorbox{backgreen}{48.7} & 55.4 & 26.4 \\
\diamondsymbol Qwen2.5-VL-Instruct-32B & 44.9 & 53.6 & 17.6 \\
\bottomrule[1.5pt]
\end{tabular}}
\vspace{-0.5cm}
\end{table}

\subsection{Error Analysis}
Diving into specific failure cases, we observe that existing models often exhibit deficiencies in width-oriented reasoning. These limitations manifest in three primary patterns:
(1) Insufficient Exploration Space: Rather than systematically examining all viable solution paths, models frequently terminate their search prematurely. In the first example, the model overlooks valid transitions from certain states (missing the path from (1,0)), resulting in incomplete enumeration during dynamic programming.
(2) Inadequate Constraint Recognition: Explicit and implicit problem constraints that should naturally prune invalid solution branches are often disregarded (\textit{e.g.}, the non-degenerate triangle constraint in the second examples).
(3) Deficient Backtracking and State Memory: Maintaining coherent solution states while cross-referencing previously explored branches proves challenging. This manifests as double-counting errors, where the model loses track of its exploration history and cannot recognize when it has already considered certain solution paths, leading to redundant and inflated final answers.

\section{Conclusion}
In this work, we introduced Think360$^\circ$, a multimodal benchmark designed to comprehensively evaluate the reasoning capabilities of MLLMs with an explicit focus on reasoning \texttt{\textbf{width}} apart from well-studied reasoning \texttt{depth}. 
Different from most previous benchmarks that primarily focus on sequential reasoning chains, Think360$^\circ$ aims to measure models' ability to \textbf{think wide} with necessary cognitive skills (\textit{e.g.}, trial-and-error, branch-and-bound, divide-and-conquer and hypothesize-and-test). 
With an elaborate workflow for benchmark construction, we collect over 1200 multimodal cases from diverse sources and perform evaluation across \textbf{12} major model series, encompassing more than \textbf{30} different MLLMs. 
Additionally, we proposed tailored tree-of-thought evaluation protocol and performed qualitative analysis of key failure patterns to provide potential directions for future research.

\section*{Acknowledgment}
This work is supported in part by the New Generation Artificial Intelligence-National Science and Technology Major Project (No. 2025ZD0123501), Beijing Natural Science Foundation (L257008, 4252054), National Natural Science Foundation of China (Grant No. 62576342, 62425606, 62550062, 32341009).

{
    \small
    \bibliographystyle{ieeenat_fullname}
    \bibliography{main}
}

\maketitlesupplementary
\setcounter{page}{1}

\renewcommand\thefigure{S\arabic{figure}}
\renewcommand\thetable{S\arabic{table}}  
\renewcommand\theequation{S\arabic{equation}}
\setcounter{equation}{0}
\setcounter{table}{0}
\setcounter{figure}{0}
\setcounter{section}{0}
\renewcommand\thesection{\Alph{section}}

\section*{Appendix}
\section{More Experiment Settings}

All experiments are conducted on A800 GPUs. Additionally, we list the maximum output length settings for different models in Table~\ref{tab:max_response_length}.

\begin{table}[!ht] 
\centering 
\caption{Maximum response length settings for different models. Models with $^{\dagger}$ are evaluated using \texttt{vLLM} (Qwen series, MiMo, Kimi, LLama, and GLM-V) or \texttt{LMDeploy} (InternVL series).}
\resizebox{0.5\textwidth}{!}{
\begin{tabular}{l|c}
\toprule[1.5pt]
\textbf{Model} & \textbf{Max Response Length} \\
\midrule
\multicolumn{2}{c}{\clubsymbol \textit{Closed-source MLLMs}} \\
\midrule
\clubsymbol GPT-4o & 16384 \\
\clubsymbol GPT-4v-preview & 16384 \\
\clubsymbol GPT-4.1 & 32768 \\
\clubsymbol o1 & 100000 \\
\clubsymbol o3 & 100000 \\
\clubsymbol o4-mini & 100000 \\
\clubsymbol Gemini-2.0-flash & 8192 \\
\clubsymbol Gemini-2.0-flash-thinking-exp-01-21 & 8192 \\
\clubsymbol Gemini-2.5-flash-thinking & 65536 \\
\clubsymbol Gemini-2.5-pro & 65536 \\
\clubsymbol Doubao-1-5-vision-pro-250328 & 16384 \\
\clubsymbol Doubao-1.5-thinking-vision-pro-250428 & 16384 \\
\clubsymbol Doubao-1.5-thinking-vision-pro-250428-nothinking & 16384 \\
\clubsymbol Claude-4-Opus-20250514 & 32000 \\
\clubsymbol Claude-4-Opus-20250514-Thinking & 32000 \\
\clubsymbol Claude-3.7-Sonnet-Thinking & 64000 \\
\clubsymbol Claude-4-Sonnet & 64000 \\
\clubsymbol Grok-2-vision-1212 & 16384 \\
\midrule
\multicolumn{2}{c}{\textit{\diamondsymbol Open-source MLLMs}} \\
\midrule
\diamondsymbol LLaVA-Onevision-7B & 8192 \\
\diamondsymbol Llama-3.2-Vision-Instruct-11B$^{\dagger}$ & 32768 \\
\diamondsymbol GLM-4.1V-Thinking-9B$^{\dagger}$ & 32768 \\
\diamondsymbol Kimi-VL-Instruct$^{\dagger}$ & 32768 \\
\diamondsymbol Kimi-VL-Thinking$^{\dagger}$ & 65536 \\
\diamondsymbol InternVL-2.5-8B$^{\dagger}$ & 8192 \\
\diamondsymbol InternVL-3-8B$^{\dagger}$ & 16384 \\
\diamondsymbol InternVL-3-14B$^{\dagger}$ & 16384 \\
\diamondsymbol InternVL-3.5-8B$^{\dagger}$ & 16384 \\
\diamondsymbol Bee-RL-7B$^{\dagger}$ & 16384 \\
\diamondsymbol MiMo-VL-RL-7B$^{\dagger}$ & 32768 \\
\diamondsymbol MiMo-VL-SFT-7B$^{\dagger}$ & 32768 \\
\diamondsymbol Qwen2.5-VL-Instruct-7B$^{\dagger}$ & 8192 \\
\diamondsymbol Qwen2.5-VL-Instruct-32B & 8192 \\
\diamondsymbol Qwen2.5-VL-Instruct-72B & 8192 \\
\bottomrule[1.5pt]
\end{tabular}}
\label{tab:max_response_length}
\end{table}

\section{Demonstration of the Scaling of Reasoning Depth \& Width}

\begin{figure*}[ht]
    \centering
    \includegraphics[width=1\linewidth]{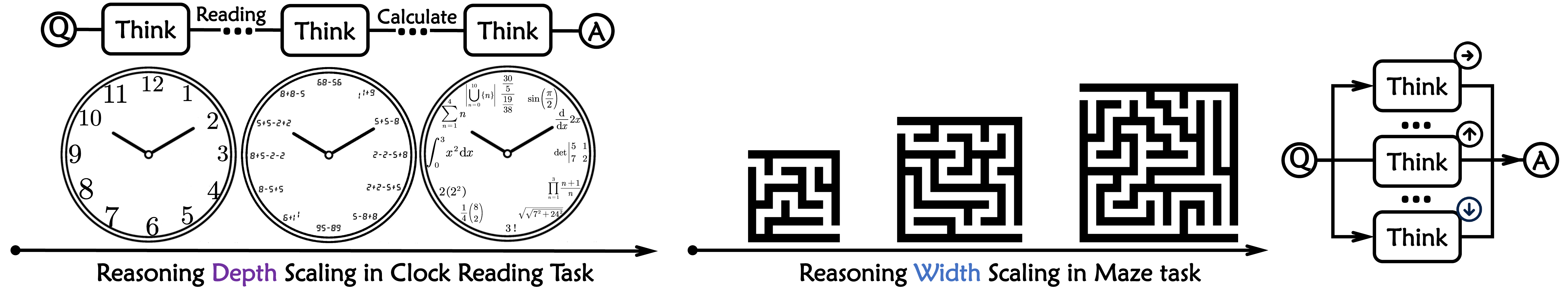}
    \caption{Demonstration of the Scaling of Reasoning Depth \& Width}
    \label{fig:depth_width_scaling_comparison}
\end{figure*}

To illustrate how reasoning depth and width scale with problem complexity, we present two representative examples in Fig.\ref{fig:depth_width_scaling_comparison}.

\noindent\textbf{Reasoning Depth.} The left demonstrates depth scaling through a clock time-reading task with progressively complex dial markings. In the simplest version, clock positions are directly labeled with numerical values, requiring only visual recognition. As complexity increases, markings transition to arithmetic expressions (addition, subtraction, multiplication, division), then to advanced mathematical operations (integrals, determinants). This progression systematically extends the sequential reasoning chain: models must parse mathematical notation, execute calculations, map results to clock positions, and determine the time—each step dependent on its predecessor. Thus, reasoning depth scales with the computational complexity embedded in visual elements.

\noindent\textbf{Reasoning Width.} The right demonstrates width scaling through maze navigation with increasing grid sizes. As the maze expands from small to medium to large configurations, the exploration space grows exponentially. Larger mazes introduce more branching points, dead ends, and alternative routes requiring parallel consideration. Models must explore multiple candidate paths simultaneously, evaluate their viability, and backtrack when necessary. The reasoning width scales directly with spatial complexity and the number of viable paths requiring concurrent evaluation.

\section{Error Analysis}
\begin{figure}[ht]
  \centering
  \begin{subfigure}{0.48\columnwidth}
    \centering
    \includegraphics[width=\linewidth]{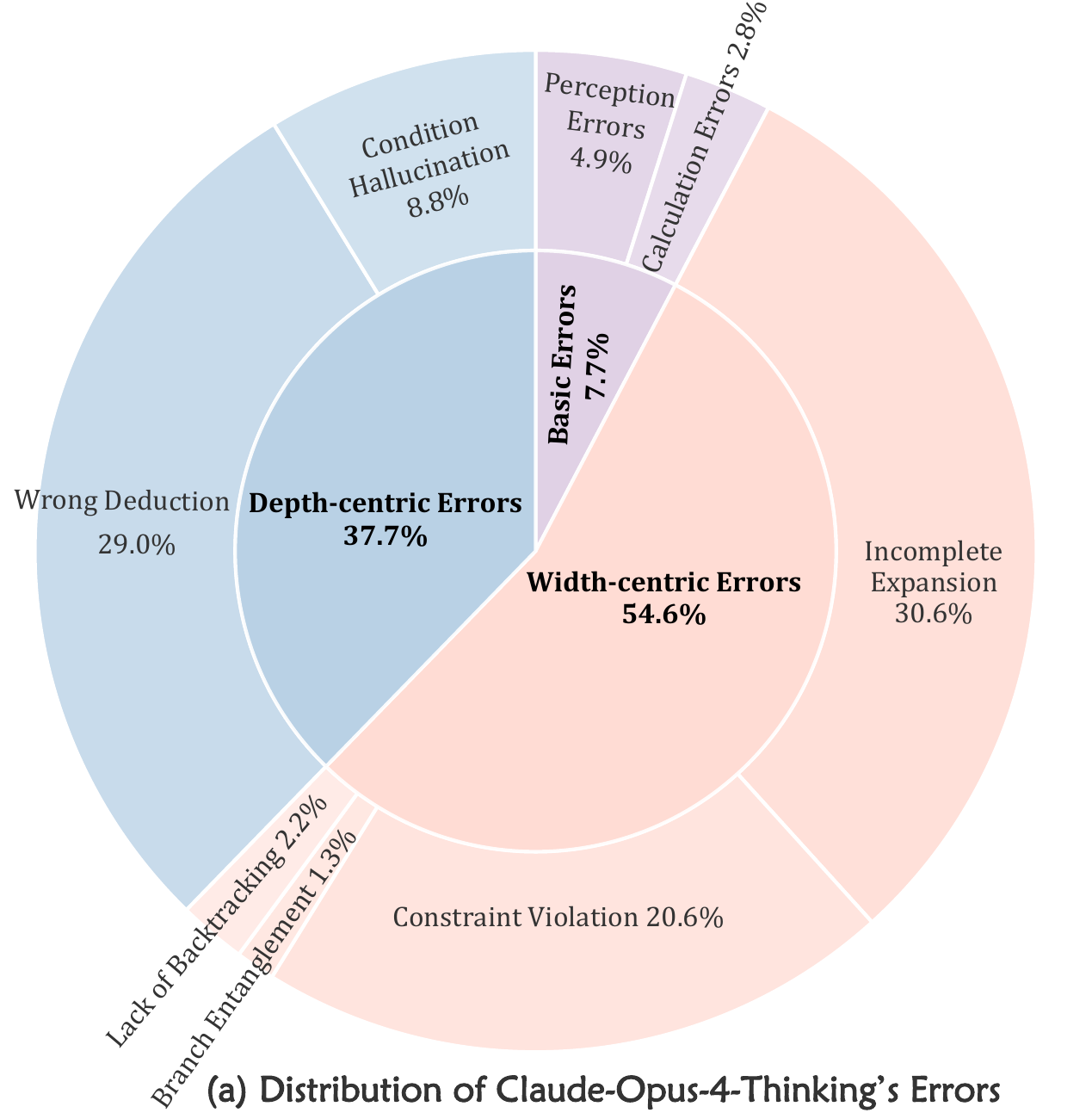}
    \label{fig:error_pie_claude_opus4_thinking}
  \end{subfigure}
  \hspace{1pt}
  \begin{subfigure}{0.48\columnwidth}
    \centering
    \includegraphics[width=\linewidth]{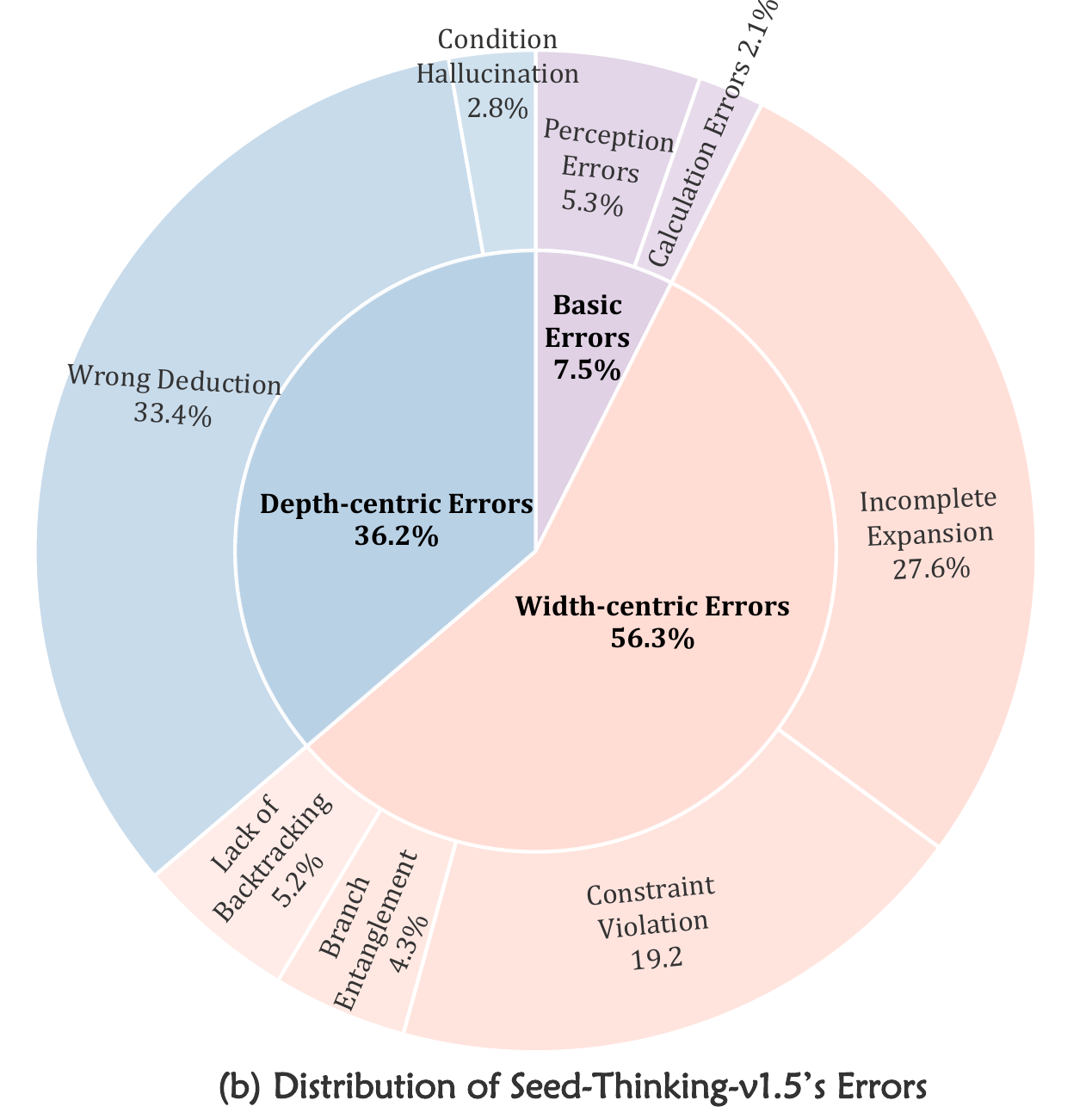}
    \label{fig:error_pie_seed}
  \end{subfigure}
 \vspace{-0.2cm}
  \caption{Distribution of error categories for Claude-Opus-4-Thinking (left) and Seed-Thinking-v1.5 (right). The pie charts illustrate that width-centric errors (e.g., incomplete branch expansion and ineffective pruning) dominate the failure modes across different architectures, substantiating that current models are primarily bottlenecked by multi-path reasoning rather than single-step depth.}
  \vspace{-0.5cm}
  \label{fig:error_pie}
\end{figure}

We conducted a fine-grained error diagnosis on two advanced thinking models: Doubao-1.5-Thinking-Vision-Pro and Claude-Opus-4-Thinking. As shown in Fig.\ref{fig:error_pie}, our analysis reveals that \textbf{width-centric errors consistently dominate}, accounting for 56\% (Doubao-1.5) and 55\% (Claude-Opus-4) of all failures. These significantly outweigh both depth-centric errors (36\%-38\%) and basic perception or calculation mistakes ($<$8\%).

A closer examination of these failure modes highlights the structural limitations driving these errors. The most prevalent issue across both models is \textit{Incomplete Branch Expansion} (accounting for $\sim$30\% of total errors). This indicates that the models arbitrarily collapse multi-path problems into a single path, failing to enumerate alternative branches (often resulting in tunnel vision). Furthermore, \textit{Ineffective Pruning} contributes to $\sim$20\% of errors, demonstrating a systematic failure to verify candidate solutions against global constraints.

\section{Prompts for Response \& Caption Generation}
In this section, we present the details of the prompts used for model evaluation and caption generation. Specifically, we employ two different prompting strategies to obtain model responses: one directly requests answers to questions, while the other first requires step-by-step reasoning before providing the final answer. The two prompting strategies are shown below:

\begin{prompt}
    \textbf{Direct Prompt:} Please answer this question.

    \textbf{CoT Prompt:} Please first think about this question step by step, and then output the final answer.
\end{prompt}

The caption generation prompt is designed to create highly detailed and accurate descriptions of visual content that serves as a bridge between visual and textual modalities, ensuring that all critical visual information required for mathematical and logical reasoning is preserved in the textual description.

Specially, we emphasizes four key principles: completeness ensures no essential visual elements are omitted; precision requires the use of exact mathematical terminology and notation; comprehensiveness mandates inclusion of every detail necessary for understanding the complete visual context; and clarity ensures logical organization of information to facilitate subsequent reasoning tasks.

\begin{prompt}
    \textbf{Caption Prompt:} You are an expert mathematical and logical reasoning analyst. 

Create an image caption so detailed and accurate that another model could reconstruct all essential visual information needed for reasoning, using only your description.

Critical Guidelines:
\begin{itemize}
    \item \textbf{Completeness}: Describe objectivel ONLY what is visually present. Do not infer, solve, interpret, or add information not explicitly shown
    \item \textbf{Precision}: Use exact mathematical terminology and standard notation
    \item \textbf{Comprehensiveness}: Include every detail necessary for another model to understand the complete visual context
    \item \textbf{Clarity}: Organize information logically to enable effective reasoning by subsequent models.
\end{itemize}

\end{prompt}

\section{Question Distribution Analysis}
\begin{figure}[ht]
    \centering
    \includegraphics[width=1\linewidth]{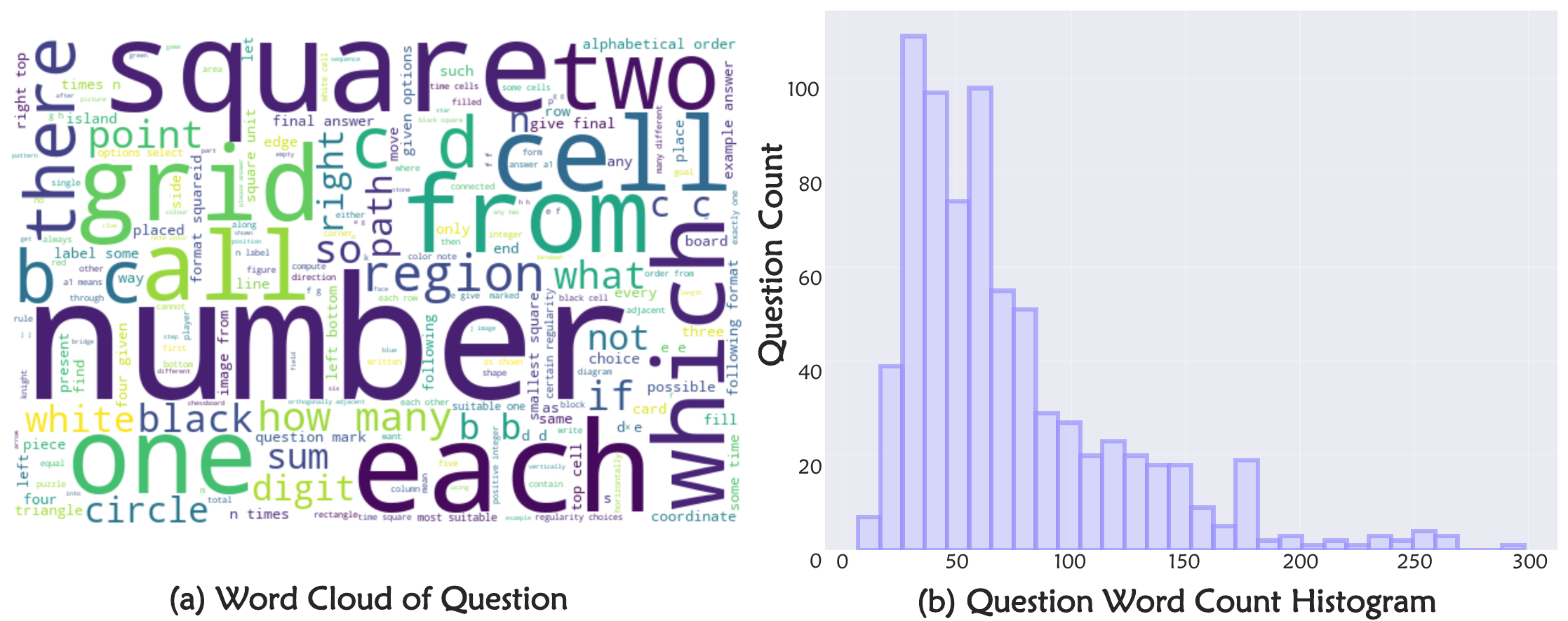}
    \caption{Word Cloud and Length Distribution Histogram of Questions in Think360.}
    \label{fig:question_wordcloud_and_length}
\end{figure}

As shown in Fig.\ref{fig:question_wordcloud_and_length}, the word cloud reveals prominent terms such as "square", "grid", "cell", and "region" indicating strong emphasis on spatial reasoning and "number", "which", "all", "how many" and "times" suggesting the demand for systematic exploration and constraint satisfaction tasks that align with our benchmark's focus on reasoning depth and width.

The distribution of question lengths reveals an average of 73.72 words, with a median of 60. What's more, it exhibits a right-skewed pattern, with the majority of questions (48\%) falling within 30-69 words. Notably, 163 questions exceed 100 words, with the longest reaching 298 words, indicating significant variation in problem context length and complexity across the dataset.

\section{Detailed Workflow for Game-based and Proof-based Question}

This section provides detailed workflows for processing two challenging question types: game-based and proof-based problems. These categories require specialized handling due to their unique characteristics—game-based problems lack explicit question-answer pairs, while proof-based problems contain non-verifiable reasoning processes that complicate objective evaluation.

\subsection{Game-based Question Processing}

\begin{figure*}[ht]
    \centering
    \includegraphics[width=1\linewidth]{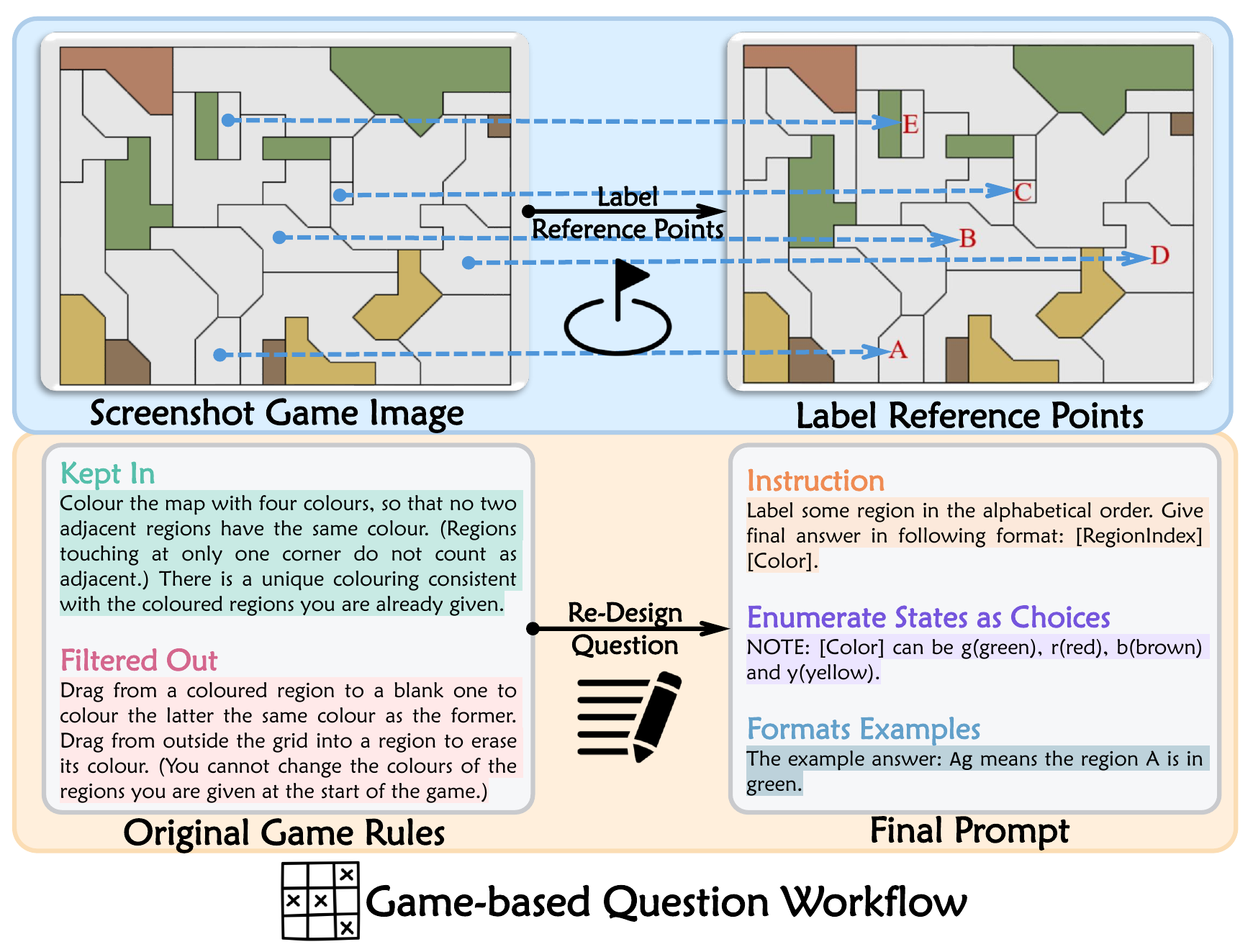}
    \caption{Workflow for Game-based Questions}
    \label{fig:workflow_game}
\end{figure*}

Game-based problems originate from interactive online puzzle games that present visual challenges without predefined questions or explicit answers. As shown in Fig.\ref{fig:workflow_game}, the processing workflow involves three key stages:

\noindent\textbf{Stage 1: Image Preparation.} We capture the initial game state screenshot that contains sufficient conditions for solving the puzzle. This image serves as the primary visual input, preserving all relevant spatial relationships and constraints. To facilitate unambiguous reference to specific regions or positions, we overlay alphabetical labels (A, B, C, etc.) onto the image, creating clearly identifiable reference points.

\noindent\textbf{Stage 2: State Enumeration and Question Design.} We systematically enumerate all possible states or values for each labeled position in the puzzle. For example, in a map coloring game, we identify the finite set of colors (e.g., green, red, brown, yellow) that can be assigned to each region. We then design questions that reference specific labeled positions (e.g., "What color is region A?"), transforming the interactive game into a well-defined question-answer format. 

\noindent\textbf{Stage 3: Format Standardization.} To ensure consistent response formats, we incorporate explicit instructions and examples directly into the problem statement. This includes specifying the answer format (e.g., "Give final answer in following format: [RegionIndex][Color]"), providing notation explanations (e.g., "g(green), r(red), b(brown), y(yellow)"), and including concrete examples (e.g., "Ag means region A is green"). This standardization enables objective verification of model responses.

Throughout this process, we filter out game instructions and UI elements that are irrelevant to the core reasoning task, retaining only the essential puzzle constraints and the designed question.

\subsection{Proof-based Question Processing}

\begin{figure*}[ht]
    \centering
    \includegraphics[width=1\linewidth]{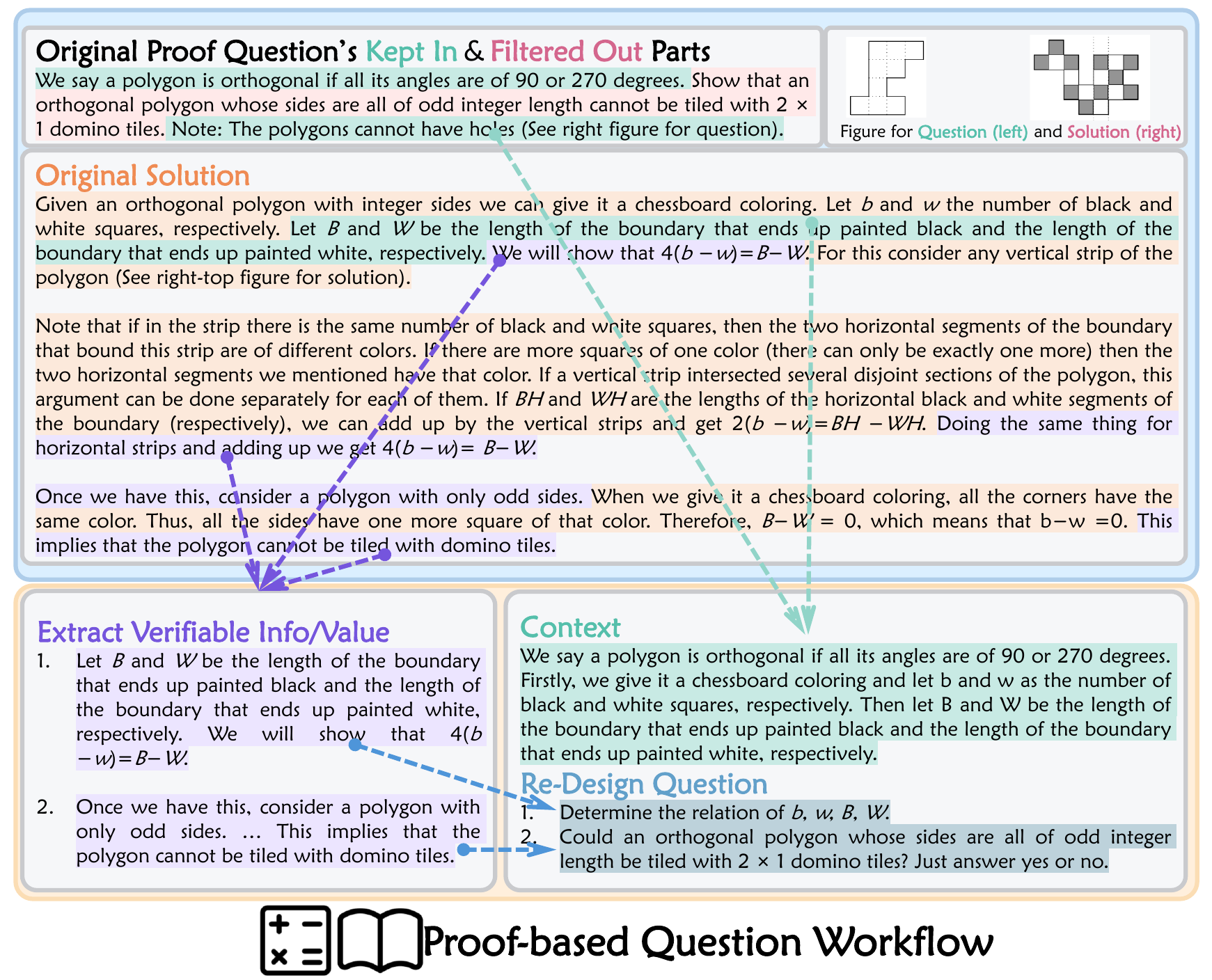}
    \caption{Workflow for Proof-based Questions}
    \label{fig:workflow_proof}
\end{figure*}

Proof-based problems from competition and textbook sources typically require demonstrating mathematical statements through logical arguments (see Fig.\ref{fig:workflow_proof}). Since complete proofs are difficult to verify objectively and may not align with our focus on visual reasoning, we adopt a redesign strategy:

\noindent\textbf{Stage 1: Proof Structure Analysis.} We carefully analyze the original proof to identify key intermediate results, numerical relationships, or specific conclusions that are objectively verifiable. For instance, in a proof showing that certain polygons cannot be tiled by dominoes, we extract intermediate results like "the relation between boundary lengths and square counts satisfies 4(b-w) = B-W" and the final conclusion "the polygon cannot be tiled."

\noindent\textbf{Stage 2: Context Extraction.} We extract the essential context from the original problem statement and proof setup that is necessary for understanding the redesigned questions. This includes definitions (e.g., "a polygon is orthogonal if all angles are 90° or 270°"), notation (e.g., "b and w denote black and white square counts"), and setup procedures (e.g., "give the polygon a chessboard coloring"). This context is incorporated into the new problem statement to ensure self-contained questions.

\noindent\textbf{Stage 3: Question Redesign.} Based on the extracted verifiable information, we redesign questions that test understanding of key proof steps or conclusions without requiring the complete proof. For example, instead of "Show that the polygon cannot be tiled," we ask "Determine the relation of b, w, B, W" or "Could such a polygon be tiled? Answer yes or no." These redesigned questions maintain mathematical rigor while enabling objective answer verification.

\noindent\textbf{Stage 4: Quality Control.} We filter out portions of the original proof process that cannot be reliably verified or that do not contribute to visual reasoning assessment. Only the redesigned questions with clear, verifiable answers are retained in the final benchmark.

This workflow transforms proof-oriented problems into evaluation-friendly formats while preserving the core mathematical reasoning required, ensuring that our benchmark remains focused on verifiable visual reasoning rather than unconstrained proof generation.

\section{More Details for ToT-Eval}
In the stage of tree construction, each node preserves the original wording and is classified by step type (e.g., calculation, deduction, conclusion). For solutions consisting only of a final answer, we create a single root node to maintain structural consistency.
\begin{prompt}
    \textbf{ToT Extraction Prompt:} You are an expert in decomposing mathematical reasoning into tree structures. Extract key reasoning steps from solutions into a hierarchical tree where depth represents sequential reasoning steps that depend on previous conclusions, and breadth captures parallel exploration of different possibilities at the same depth level.
    
Critical Guidelines:
\begin{itemize}
    \item \textbf{Verbatim Extraction}: Each node content must be directly extracted from the original solution text, preserving the original wording without paraphrasing
    \item \textbf{Critical Steps Only}: Focus on major logical leaps, calculations, and key deductions rather than simple listings or obvious observations. Keep complete calculation steps as one node (e.g., "ans = 4 + 7 = 11" should not be split)
    \item \textbf{Tree Structure}: Use node ID format \{depth\}.\{sequence\}, where child depth = parent depth + 1. Depth 1 nodes must have parent = None (root nodes), and nodes at the same depth are siblings, not parent-child
    \item \textbf{Special Case}: For solutions containing only a simple final answer (e.g., "D", "42"), create a single root node with parent = None to maintain structural consistency
\end{itemize}
\end{prompt}

After constructing the reasoning tree and judging each node's correctness, we compute two complementary metrics to quantify reasoning depth and breadth. ToT-Depth measures the quality of the deepest reasoning chains by averaging the accuracy along all paths from root to maximum-depth leaves, thereby rewarding long, correct chains while penalizing early logical breakdowns. ToT-Width measures the quality of parallel reasoning by averaging the accuracy across sibling groups at each depth level, crediting models that successfully explore multiple valid branches. Together, these process-based metrics provide fine-grained assessment of long chain-of-thought responses beyond simple outcome-based accuracy.
\begin{prompt}
    \textbf{ToT Judgement Prompt:} You are an expert in evaluating mathematical and logical reasoning steps. Judge the correctness of a single reasoning step within a larger reasoning tree, considering the problem context, reference answer, parent nodes for context, and the image if relevant.
    
Evaluation Criteria:
\begin{itemize}
    \item \textbf{Correctness}: Is the reasoning in this step logically sound and factually correct?
    \item \textbf{Validity}: Does it follow properly from the parent nodes?
    \item \textbf{Accuracy}: For calculations, are the results mathematically correct?
    \item \textbf{Relevance}: Does it contribute meaningfully to solving the problem?
    \item \textbf{Final Answer Check}: If this is the final answer node, does it match the reference answer?
\end{itemize}

Output: Respond with only "True" (if correct) or "False" (if incorrect/flawed). Note that intermediate steps can be correct even if the final answer is wrong, and conversely, a step can be flawed even if it leads to the correct final answer.
\end{prompt}

\noindent\textbf{ToT-Depth} measures the quality of the deepest reasoning chains. We first identify all leaf nodes at the maximum depth $d_{\max}$, then trace back from each leaf to the root to form complete reasoning paths. For each path $P_i$ containing nodes $\{n_1, n_2, \ldots, n_{|P_i|}\}$, we calculate its accuracy as the proportion of correct nodes. The final depth score is the average accuracy across all deepest paths:
\begin{equation}
\text{ToT-Depth} = \frac{1}{|P|} \sum_{i=1}^{|P|} \frac{\sum_{n \in P_i} \mathbb{1}[\text{correct}(n)]}{|P_i|}
\end{equation}
where $P$ is the set of all paths from root to maximum-depth leaves, and $\mathbb{1}[\text{correct}(n)]$ indicates whether node $n$ is judged correct.

\noindent\textbf{ToT-Width} measures the quality of parallel reasoning exploration. We group all nodes by their parent, identifying sibling groups that represent alternative reasoning branches at the same depth level. For each parent node $p$ with children $C_p = \{c_1, c_2, \ldots, c_{|C_p|}\}$, we compute the accuracy of this sibling group. The width score is the average accuracy across all such groups:
\begin{equation}
\text{ToT-Width} = \frac{1}{|G|} \sum_{p \in G} \frac{\sum_{c \in C_p} \mathbb{1}[\text{correct}(c)]}{|C_p|}
\end{equation}
where $G$ is the set of all parent nodes that have at least one child. These metrics together provide a comprehensive view of both the vertical depth and horizontal breadth of the model's reasoning capabilities.

\section{More Experiments Results}
In this section, we provide more experiments results.

\begin{figure}[t]
    \centering
    \includegraphics[width=1\linewidth]{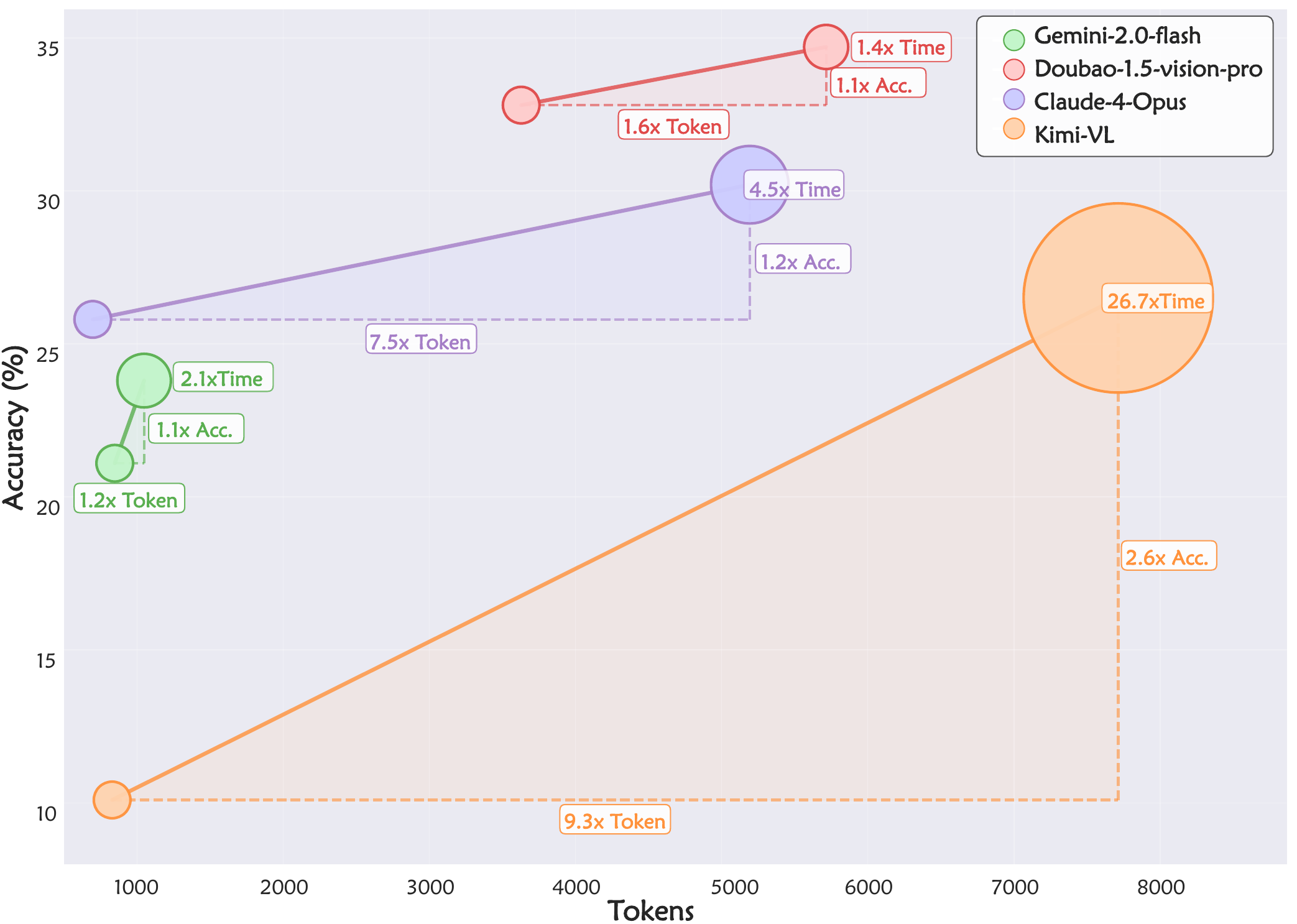}
    \caption{Thinking Mode Ablation. The x-axis shows accuracy improvement, and the y-axis shows token increase. Bubble size visualizes the time cost expansion. For each model, we fix the radius of the non-thinking circle to 1 and scale the radius of the thinking circle by the empirical multiplier of inference time (×Time). }
    \vspace{-0.5cm}
    \label{fig:thinking_mode}
\end{figure}

\begin{figure*}[!ht]
    \centering
    \begin{subfigure}{\linewidth}
        \centering
        \includegraphics[width=\linewidth]{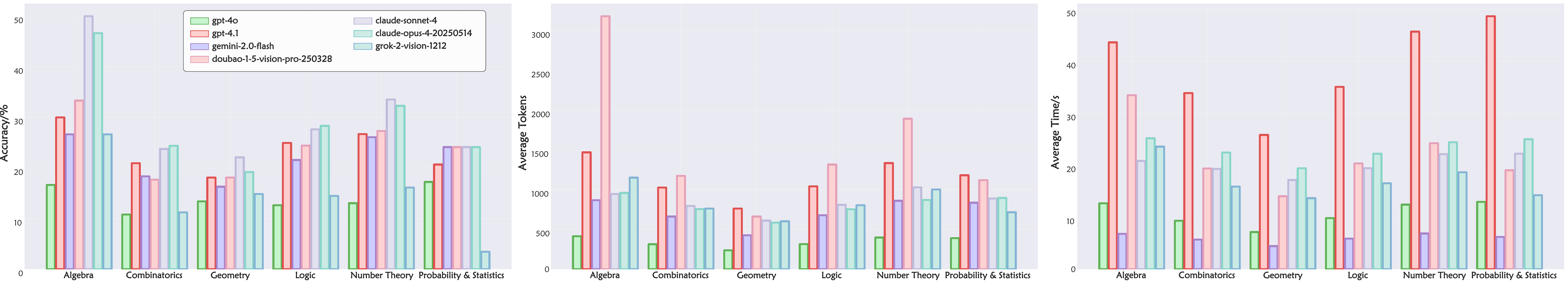}
        \caption{System-1 MLLMs with fast and single-pass reasoning (average time$<50s$, token$<3000$)}
        \label{fig:sub_category_comparison_nothinking}
    \end{subfigure}
    \vspace{0.1cm}
    \begin{subfigure}{\linewidth}
        \centering
        \includegraphics[width=\linewidth]{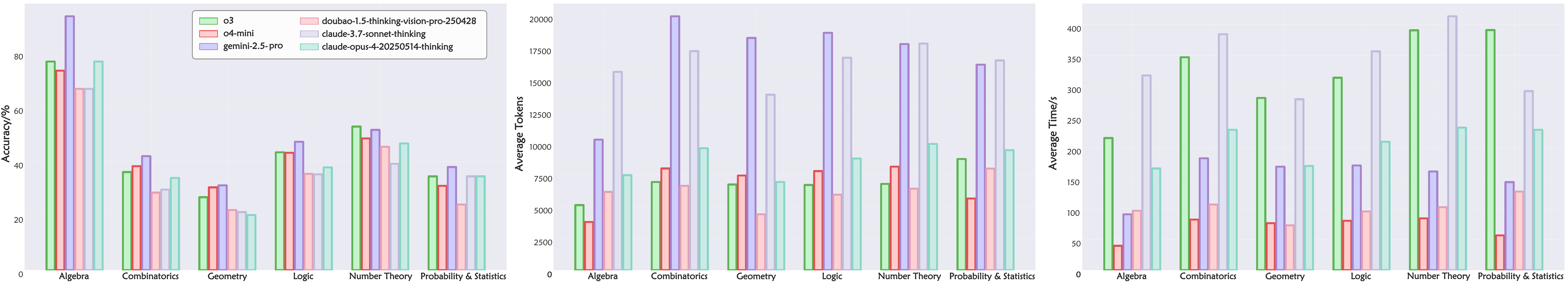}
        \caption{System-2 MLLMs with slow and iterative long CoT reasoning (average time$>50s$, token$>3000$).}
        \label{fig:sub_category_comparison_thinking}
    \end{subfigure}
    \caption{Reasoning performances comparison across different question types. To account for the substantial disparities in reasoning time and token usage across models, we categorize them using predefined time/token thresholds to better highlight the performance profiles of system-1 and system-2 models. Please zoom in for a better view.} 
    \label{fig:sub_category_comparison}
\end{figure*}

\noindent\textbf{Performance Patterns Across Subject:} 
As shown in Fig.~\ref{fig:sub_category_comparison}, models demonstrate above-average performance in Algebra and Number Theory subsets, but underperform in Combinatorics, Geometry, and Probability \& Statistics. This gap suggests that current MLLMs are more comfortable with problems that can be reduced to relatively direct symbolic manipulation or formula-based computation, while they struggle with tasks that require exploring large combinatorial spaces, handling spatial relations, or modeling uncertainty. In particular, the deficits in Combinatorics and Probability \& Statistics are consistent with the difficulty of width-oriented reasoning, where models must juggle multiple cases, scenarios, or distributions rather than follow a single dominant derivation path.

\noindent\textbf{Thinking vs. No Thinking:} 
As shown in the Fig.\ref{fig:thinking_mode}, enabling thinking mode yields markedly different trade-offs across models. Kimi-VL sits at the “heavy-thinking” extreme: it gains 2.6× accuracy but at the cost of 9.3× tokens and 26.7× inference time, making it the most expensive option in test-time scaling. Claude-4-Opus shows a milder pattern, with 1.2× accuracy, 7.5× tokens, and 4.5× time, indicating more controlled but still substantial overhead.

By contrast, Doubao-1.5-vision-pro and Gemini-2.0-flash behave like “lightweight thinking” models: they achieve around 1.1× accuracy with only 1.4×–2.1× time and 1.2×–1.6× tokens. Overall, these results suggest that some models (e.g., Kimi-VL) aggressively trade latency for gains, while others (e.g., Doubao, Gemini) offer a more balanced accuracy–efficiency compromise.

What's more, Tab.\ref{tab:cot_comparison_full} demonstrates more fine-grained influence of Chain-of-Thought prompting on the splits requiring different cognitive capabilities.

\begin{table*}[!ht] 
\centering 
\caption{Fine-grained Influence of Chain-of-Thought prompting on model performances. 
} 
\resizebox{\textwidth}{!}{
\begin{tabular}{l|c|c|c|c|c|c|c|c|c|c|c|c|c|c|c|c|c|c|c|c}
\toprule[1.5pt]
Model & \#Para. & CoT & \multicolumn{3}{c|}{\cellcolor{magenta!20}\textbf{ALL}} & \multicolumn{3}{c|}{\cellcolor{orange!20}Perceive‑and‑Comprehend} & \multicolumn{3}{c|}{\cellcolor{yellow!20}Trial-and-Error} & \multicolumn{3}{c|}{\cellcolor{pink!20}Divide‑and-Conquer} & \multicolumn{3}{c|}{\cellcolor{lime!20}Branch‑and-Bound} & \multicolumn{3}{c}{\cellcolor{cyan!20}Hypothesize-and-Test} \\
\midrule
& & & Acc./$$\%$$ & Time/$s$ & Token & Acc./$\%$ & Time/$s$ & Token & Acc./$\%$ & Time/$s$ & Token & Acc./$\%$ & Time/$s$ & Token & Acc./$\%$ & Time/$s$ & Token & Acc./$\%$ & Time/$s$ & Token \\
\midrule
\midrule
\multicolumn{21}{c}{\clubsymbol \textit{Close-source MLLMs}}\\
\midrule
\midrule
\multirow{3}{*}{\clubsymbol GPT-4o} & \multirow{3}{*}{-} & \ding{55} & 16.0 & 13.28 & 309.03 & 17.2 & 12.69 & 287.16 & 15.3 & 10.72 & 268.64 & 9.9 & 12.21 & 331.46 & 16.8 & 13.17 & 322.85 & 14.3 & 12.89 & 313.87  \\
& & \ding{51} & 16.4 & 28.41 & 388.91 & 17.5 & 27.53 & 321.81 & 15.1 & 20.88 & 314.74 & 12.2 & 33.59 & 432.74 & 16.3 & 22.56 & 458.20 & 13.8 & 28.07 & 384.97 \\
& & $\Delta$ & \up{0.4} & \up{15.1} & \up{79.9} & \up{0.3} & \up{14.8} & \up{34.6} & \down{0.2} & \up{10.2} & \up{46.1} & \up{2.3} & \up{21.4} & \up{101.3} & \down{0.5} & \up{9.4} & \up{135.4} & \down{0.5} & \up{15.2} & \up{71.1} \\
\midrule 
\multirow{3}{*}{\clubsymbol o4-mini} & \multirow{3}{*}{-} & \ding{55} & 42.1 & 84.61 & 6736.37 & 42.8 & 81.37 & 6391.21 & 34.3 & 106.52 & 8067.57 & 38.7 & 89.62 & 7460.26 & 48.0 & 76.56 & 6401.12 & 37.9 & 91.37 & 7195.80  \\
& & \ding{51} & 43.4 & 78.35 & 7079.01 & 44.6 & 75.00 & 6760.91 & 37.1 & 94.76 & 8651.76 & 37.2 & 84.31 & 7836.95 & 52.8 & 72.72 & 6071.11 & 37.8 & 82.15 & 7511.60 \\
& & $\Delta$ & \up{1.3} & \down{6.3} & \up{342.6} & \up{1.8} & \down{6.4} & \up{369.7} & \up{2.8} & \down{11.8} & \up{584.2} & \down{1.5} & \down{5.3} & \up{376.7} & \up{4.8} & \down{3.8} & \down{330.0} & \down{0.1} & \down{9.2} & \up{315.8} \\
\midrule 
\multirow{3}{*}{\clubsymbol Claude-4-Opus-20250514} & \multirow{3}{*}{-} & \ding{55} & 25.8 & 41.45 & 696.93 & 26.7 & 40.50 & 671.71 & 22.3 & 42.40 & 694.89 & 18.2 & 44.50 & 714.85 & 28.7 & 42.66 & 714.23 & 21.9 & 42.57 & 710.77  \\
& & \ding{51} & 30.4 & 19.89 & 722.09 & 32.6 & 19.31 & 685.16 & 26.6 & 20.52 & 729.83 & 19.6 & 19.94 & 727.74 & 33.3 & 20.13 & 758.95 & 25.5 & 20.11 & 731.80 \\
& & $\Delta$ & \up{4.6} & \down{21.6} & \up{25.2} & \up{5.9} & \down{21.2} & \up{13.4} & \up{4.3} & \down{21.9} & \up{34.9} & \up{1.4} & \down{24.6} & \up{12.9} & \up{4.6} & \down{22.5} & \up{44.7} & \up{3.6} & \down{22.5} & \up{21.0} \\
\midrule 
\multirow{3}{*}{\clubsymbol Claude-4-Sonnet} & \multirow{3}{*}{-} & \ding{55} & 28.2 & 18.59 & 785.22 & 29.7 & 18.04 & 747.34 & 23.2 & 18.58 & 752.49 & 20.9 & 18.79 & 830.18 & 30.9 & 18.81 & 824.35 & 24.1 & 18.93 & 796.31  \\
& & \ding{51} & 28.2 & 15.75 & 797.74 & 29.2 & 15.28 & 762.82 & 23.1 & 15.66 & 778.02 & 23.0 & 15.96 & 818.09 & 35.0 & 16.09 & 826.26 & 23.8 & 15.80 & 797.85 \\
& & $\Delta$ &\up{0.0} & \down{2.8} & \up{12.5} & \down{0.5} & \down{2.8} & \up{15.5} & \down{0.1} & \down{2.9} & \up{25.5} & \up{2.1} & \down{2.8} & \down{12.1} & \up{4.1} & \down{2.7} & \up{1.9} & \down{0.3} & \down{3.1} & \up{1.5} \\
\midrule 
\multirow{3}{*}{\clubsymbol Grok-2-vision-1212} & \multirow{3}{*}{-} & \ding{55} & 15.7 & 15.81 & 763.63 & 16.3 & 15.26 & 728.68 & 13.0 & 16.65 & 790.89 & 8.1 & 15.47 & 760.14 & 17.9 & 14.38 & 673.29 & 14.0 & 15.83 & 775.08  \\
& & \ding{51} & 17.3 & 30.35 & 764.20 & 19.6 & 27.52 & 667.04 & 14.8 & 34.94 & 773.16 & 12.8 & 39.25 & 699.59 & 9.8 & 24.35 & 734.44 & 15.5 & 29.86 & 783.83 \\
& & $\Delta$ & \up{1.60} & \up{14.54} & \up{0.57} & \up{3.30} & \up{12.26} & \down{61.64} & \up{1.80} & \up{18.29} & \down{17.73} & \up{4.70} & \up{23.78} &  \down{60.55} & \down{8.10} & \up{9.97} & \up{61.15} & \up{1.50} & \up{14.03} & \up{8.75} \\
\midrule
\midrule
\multicolumn{21}{c}{\diamondsymbol \textit{Open-source MLLMs}}\\ 
\midrule
\midrule
\multirow{3}{*}{\diamondsymbol Kimi-VL-Instruct$^{\dagger}$} & \multirow{3}{*}{16A3B} & \ding{55} & 10.1 & 39.79 & 829.45 & 11.1 & 38.14 & 750.18 & 7.7 & 49.11 & 1029.10 & 5.9 & 43.66 & 793.32 & 9.8 & 39.22 & 860.11 & 9.3 & 41.07 & 852.28  \\
& & \ding{51} & 10.9 & 30.20 & 775.96 & 12.6 & 27.94 & 679.89 & 8.1 & 35.44 & 960.65 & 5.4 & 34.09 & 815.00 & 10.9 & 28.80 & 728.97 & 10.2 & 31.52 & 814.84 \\
& & $\Delta$ & \up{0.8} & \down{9.6} & \down{53.5} & \up{1.5} & \down{10.2} & \down{70.3} & \up{0.5} & \down{13.7} & \down{68.4} & \down{0.5} & \down{9.6} & \up{21.7} & \up{1.1} & \down{10.4} & \down{131.1} & \up{0.9} & \down{9.5} & \down{37.4} \\
\midrule
\multirow{3}{*}{\diamondsymbol Qwen2.5-VL-Instruct$^{\dagger}$} & \multirow{3}{*}{7B} & \ding{55} & 11.0 & 20.31 & 788.44 & 13.0 & 18.62 & 711.00 & 8.7 & 23.43 & 924.76 & 5.8 & 21.03 & 837.41 & 10.6 & 18.05 & 692.89 & 10.3 & 20.95 & 815.10  \\
& & \ding{51} & 11.1 & 14.21 & 649.44 & 12.8 & 13.52 & 605.39 & 8.2 & 15.43 & 723.83 & 8.6 & 14.16 & 662.28 & 10.6 & 14.46 & 634.16 & 10.0 & 14.42 & 662.73 \\
& & $\Delta$ & \up{0.1} & \down{6.1} & \down{139.0} & \down{0.2} & \down{5.1} & \down{105.6} & \down{0.6} & \down{8.0} & \down{200.9} & \up{2.8} & \down{6.9} & \down{175.1} & \up{0.0} & \down{3.6} & \down{58.7} & \down{0.3} & \down{6.5} & \down{152.4} \\
\midrule
\multirow{3}{*}{\diamondsymbol Qwen2.5-VL-Instruct} & \multirow{3}{*}{72B} & \ding{55} & 16.2 & 24.20 & 615.09 & 18.2 & 23.63 & 587.27 & 14.4 & 25.87 & 652.28 & 9.0 & 24.46 & 618.13 & 13.6 & 23.65 & 608.68 & 13.9 & 24.39 & 623.97  \\
& & \ding{51} & 15.9 & 24.61 & 627.19 & 18.3 & 23.85 & 592.46 & 14.0 & 26.32 & 667.55 & 8.8 & 25.78 & 641.36 & 16.3 & 26.36 & 629.20 & 13.6 & 23.54 & 622.19 \\
& & $\Delta$ & \down{0.3} & \up{0.4} & \up{12.1} & \up{0.1} & \up{0.2} & \up{5.2} & \down{0.4} & \up{0.4} & \up{15.3} & \down{0.2} & \up{1.3} & \up{23.2} & \up{2.7} & \up{2.7} & \up{20.5} & \down{0.3} & \down{0.9} & \down{1.8} \\
\midrule
\multirow{3}{*}{\diamondsymbol InternVL-3$^{\dagger}$} & \multirow{3}{*}{8B} & \ding{55} & 13.1 & 4.67 & 379.43 & 14.2 & 4.45 & 360.55 & 12.3 & 4.98 & 405.78 & 8.6 & 5.06 & 413.24 & 15.4 & 4.81 & 391.61 & 11.4 & 4.74 & 385.42  \\
& & \ding{51} & 12.3 & 5.31 & 432.41 & 13.1 & 5.17 & 420.59 & 10.9 & 5.47 & 447.35 & 4.7 & 5.49 & 449.65 & 12.5 & 5.22 & 426.41 & 10.9 & 5.41 & 441.31 \\
& & $\Delta$ & \down{0.8} & \up{0.6} & \up{53.0} & \down{1.1} & \up{0.7} & \up{60.0} & \down{1.3} & \up{0.5} & \up{41.6} & \down{3.8} & \up{0.4} & \up{36.4} & \down{3.0} & \up{0.4} & \up{34.8} & \down{0.5} & \up{0.7} & \up{55.9}  \\
\midrule
\multirow{3}{*}{\diamondsymbol MiMo-VL-RL$^{\dagger}$} & \multirow{3}{*}{7B} & \ding{55} & 28.3 & 334.21 & 7380.78 & 29.7 & 314.61 & 6870.68 & 24.9 & 348.01 & 7761.05 & 19.1 & 394.47 & 8446.60 & 27.9 & 281.06 & 7226.50 & 24.4 & 352.59 & 7692.95 \\
& & \ding{51} & 29.9 & 345.61 & 7941.18 & 31.3 & 325.85 & 7393.75 & 25.6 & 376.58 & 8255.09 & 21.4 & 395.02 & 8772.46 & 31.4 & 306.17 & 8115.06 & 25.9 & 369.66 & 8338.44 \\
& & $\Delta$ & \up{1.6} & \up{11.4} & \up{560.4} & \up{1.6} & \up{11.2} & \up{523.1} & \up{0.7} & \up{28.6} & \up{494.0} & \up{2.3} & \up{0.5} & \up{325.9} & \up{3.5} & \up{25.1} & \up{888.6} & \up{1.4} & \up{17.1} & \up{645.5} \\
\bottomrule [1.5pt]
\end{tabular}
} 
\label{tab:cot_comparison_full} 
\end{table*}

\begin{figure*}[ht]
    \centering
    \includegraphics[width=1\linewidth]{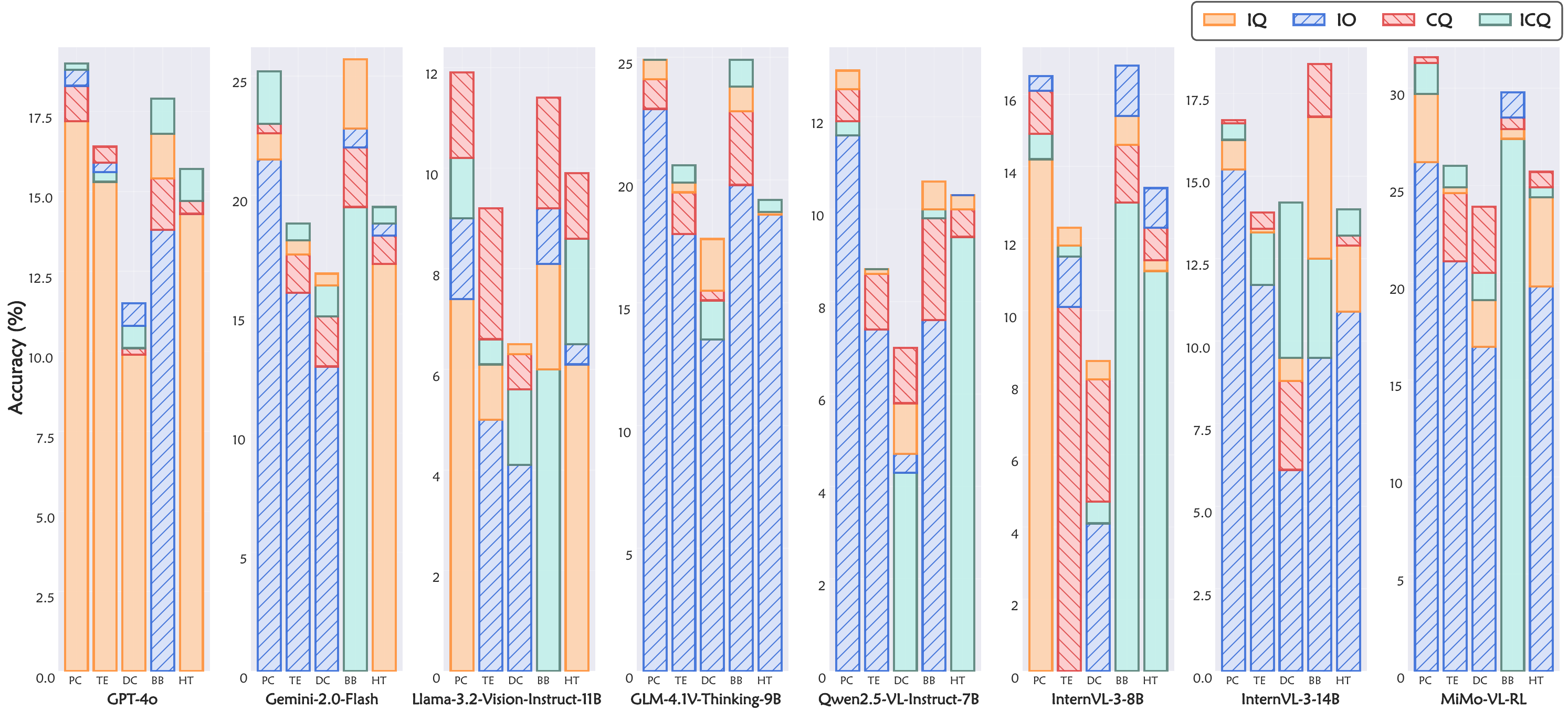}
    \caption{Ablation Study of Input Settings.}
    \vspace{-0.5cm}
    \label{fig:input_ablation}
\end{figure*}
\noindent\textbf{Image vs. Text:}
To further enrich our benchmarks and disentangle visual perception and reasoning, we design four different input settings based on the vision or language modalities. A question can be represented by four different ways: Image+Question (IQ), Image-Only (IO), Caption+Question (CQ) and Image+Question+Caption (ICQ). Specifically, Image-Only version problems are generated by adaptively overlaying the original question text onto the image according to the original image's aspect ratio. The caption with detailed descriptive information about the original image is generated by advanced o4-mini.

The ablation results in Fig.\ref{fig:input_ablation} point to a consistent pattern: Image-Only (IO) tends to be the hardest setting, underperforming Image+Question (IQ) in a clear majority of cases (60\%) and underscoring persistent challenges in visual perception within current MLLMs.
By contrast, Caption+Question (CQ) generally yields small but reliable improvements over IQ, with substantial improvements in certain models such as Llama-3.2-Vision (2.9\%) and MiMo-VL-RL (1.7\%).
Interestingly, Image+Question+Caption (ICQ) typically achieves comparable performance to CQ rather than decisively surpassing it, implying that once a high-quality caption is available, additional raw visual inputs offer limited incremental value, likely due to information redundancy or multimodal fusion overhead. 
These trends highlight two key takeaways: (i) a persistent perception bottleneck under image-only inputs, and (ii) the value of textual captions as an effective bridge between vision and language, even when the marginal benefit of reintroducing images remains modest. 

\section{More Examples}
In this section, we provide more data points in our benchmarks for intuitive understanding.

\begin{figure}[ht]
    \centering
    \includegraphics[width=1\linewidth]{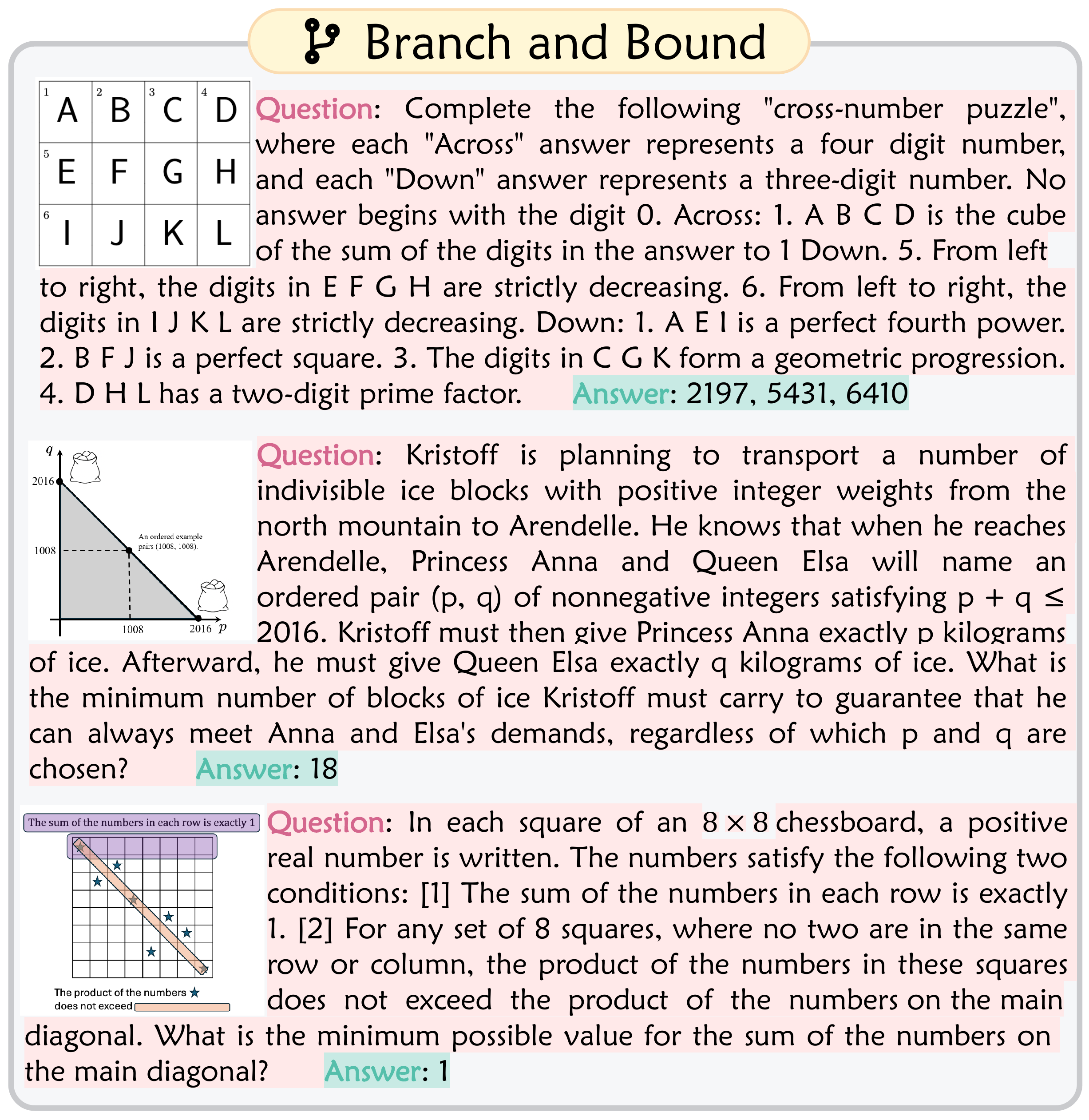}
    \caption{Examples of the Branch-and-Bound main category.}
    \label{fig:app_main_branch_and_bound}
\end{figure}

\begin{figure}[ht]
    \centering
    \includegraphics[width=1\linewidth]{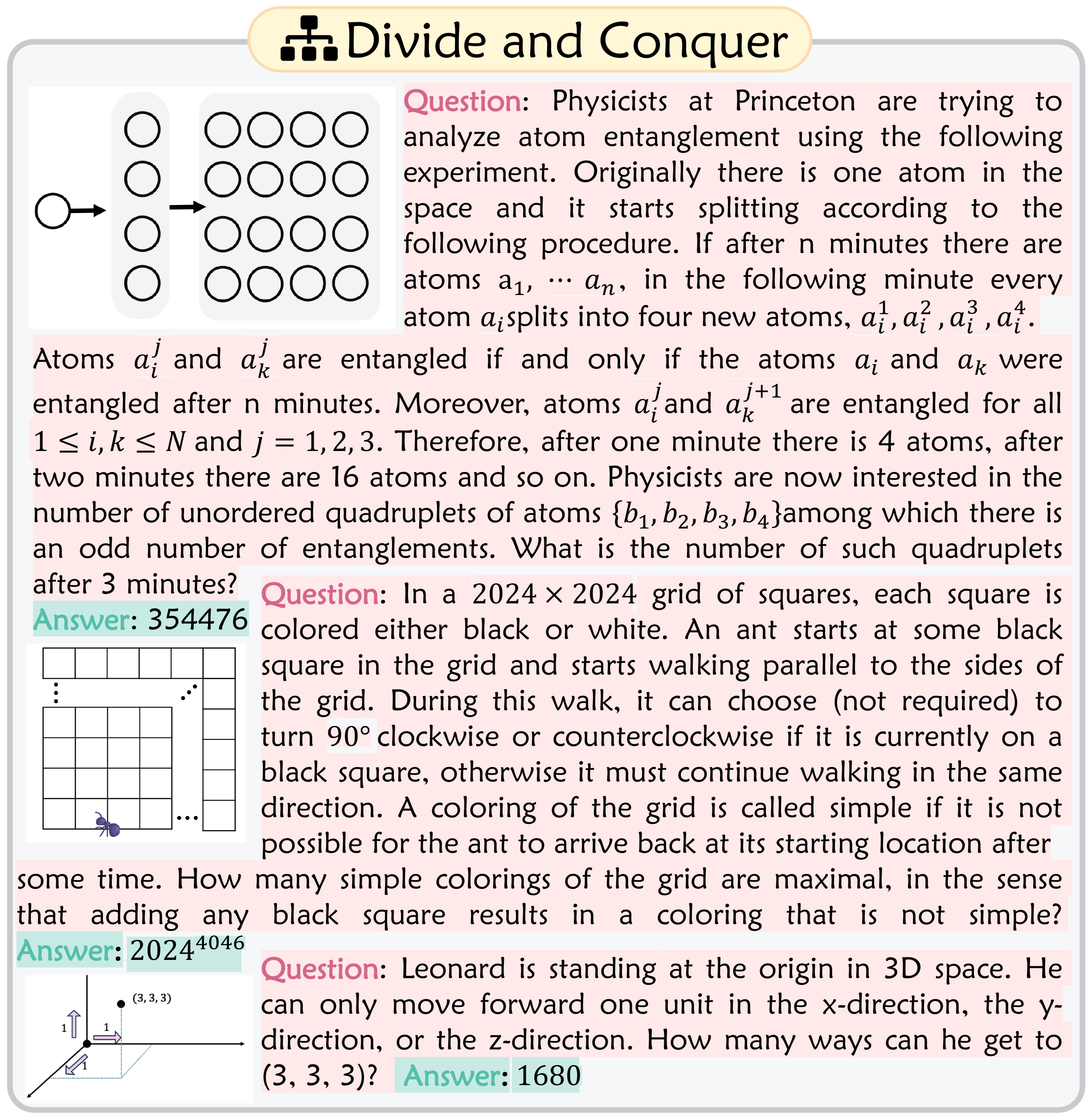}
    \caption{Examples of the Divide-and-Conquer main category.}
    \label{fig:app_main_divide_and_conquer}
\end{figure}

\begin{figure}[ht]
    \centering
    \includegraphics[width=1\linewidth]{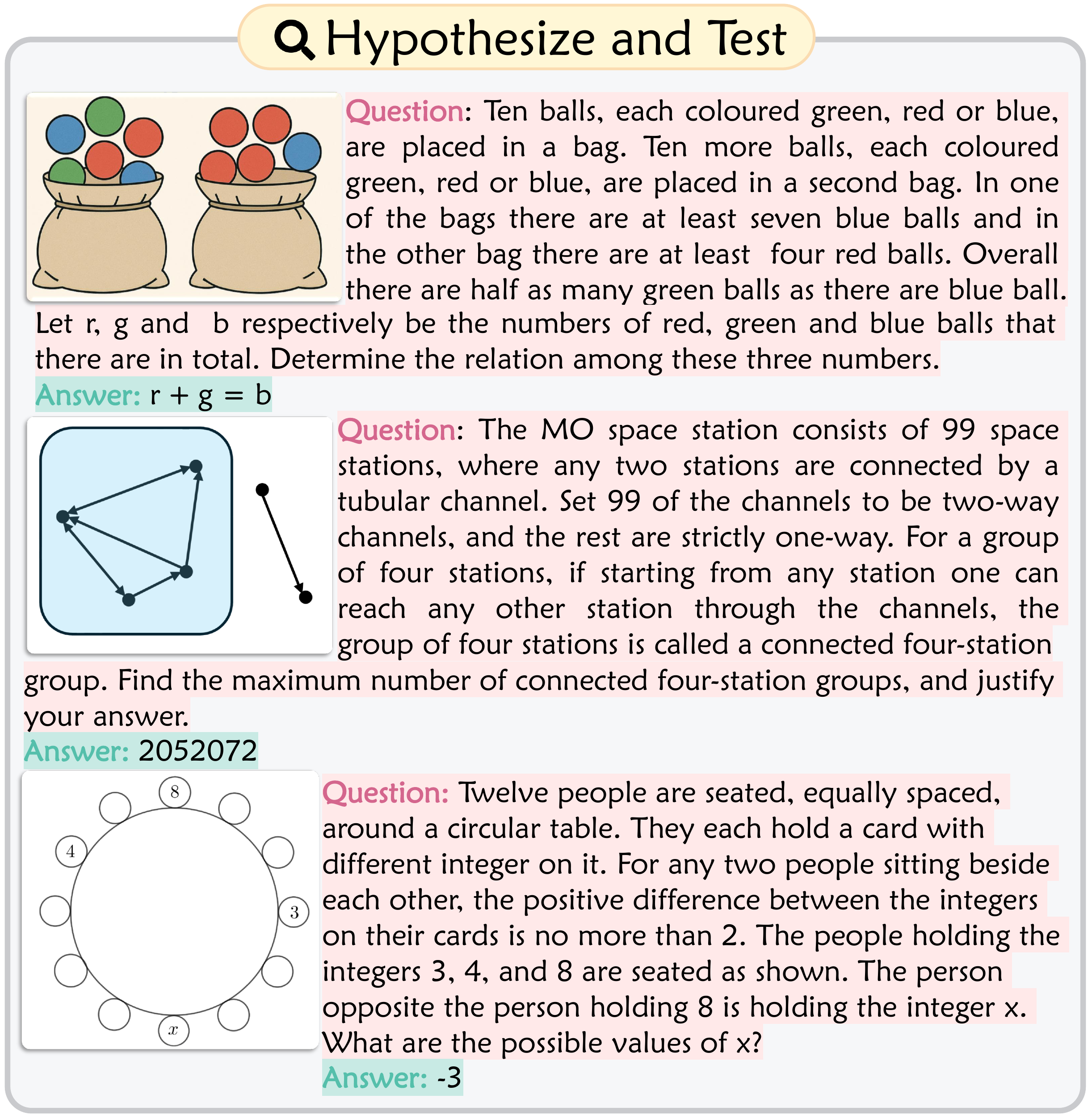}
    \caption{Examples of the Hypothesize-and-Test main category.}
    \label{fig:app_main_hypo_and_test}
\end{figure}

\begin{figure}[ht]
    \centering
    \includegraphics[width=1\linewidth]{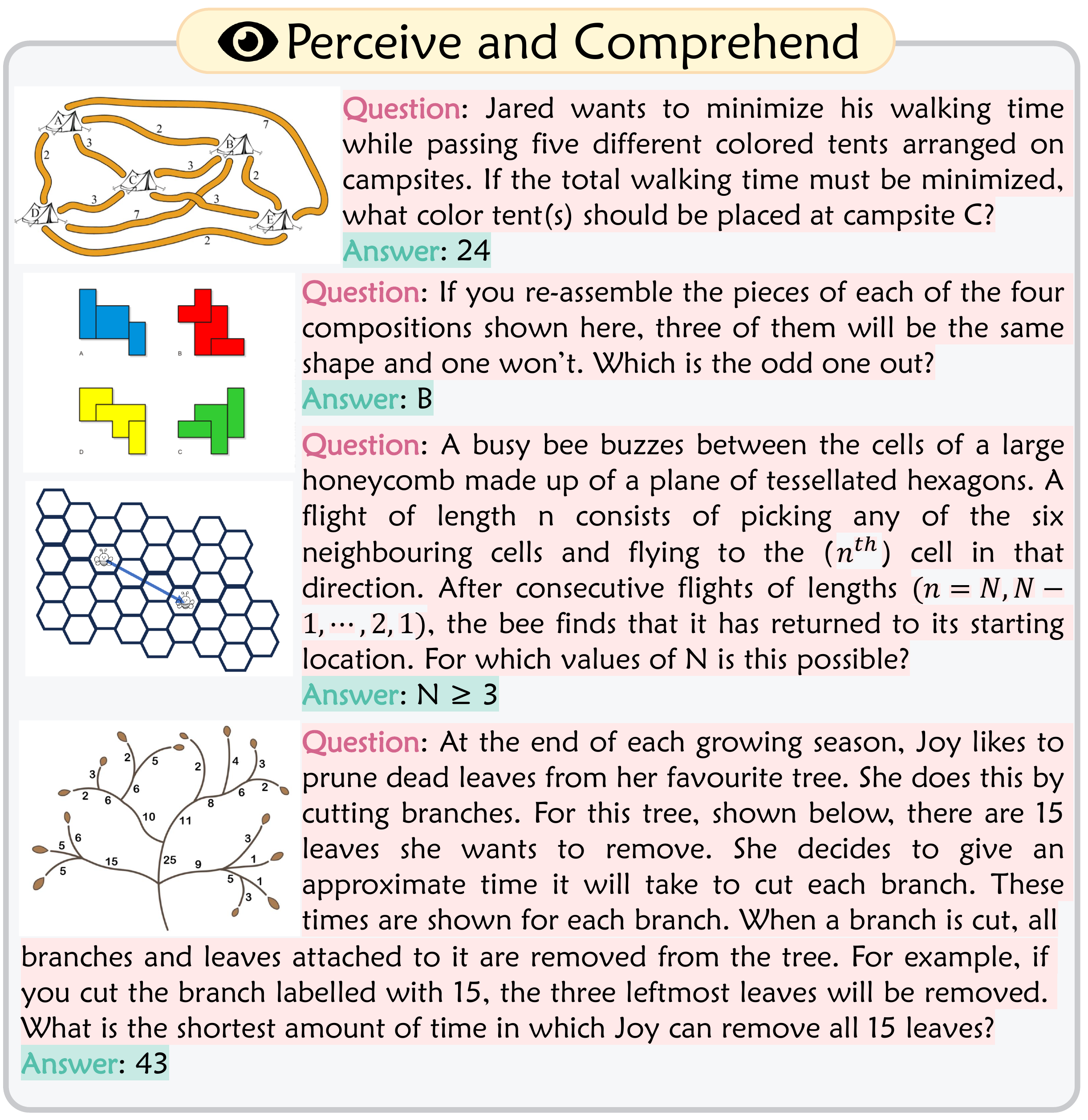}
    \caption{Examples of the Perceive-and-Comprehend main category.}
    \label{fig:app_main_perceive_and_comprehend}
\end{figure}

\begin{figure}[ht]
    \centering
    \includegraphics[width=1\linewidth]{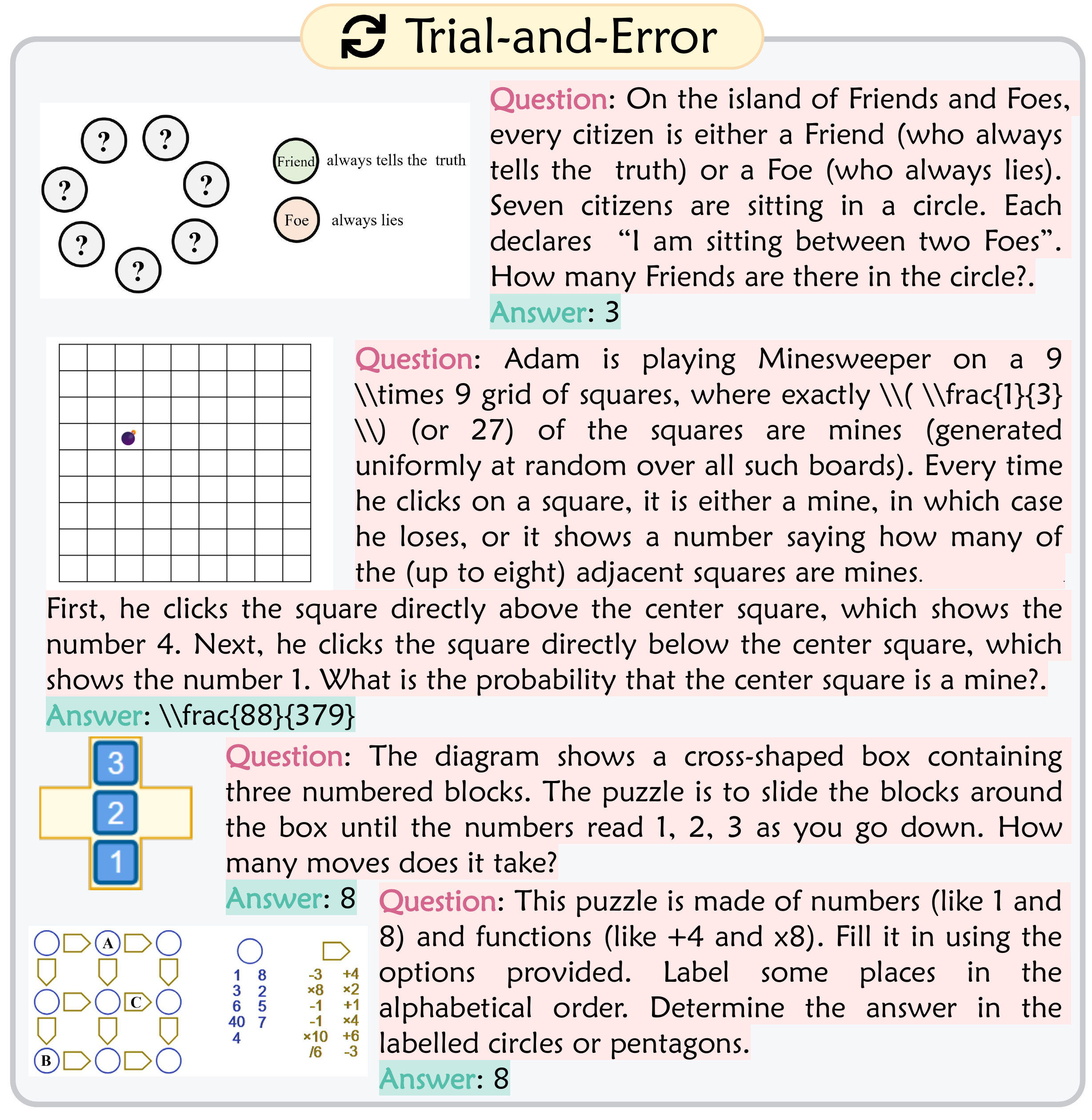}
    \caption{Examples of the Trial-and-Error main category.}
    \label{fig:app_main_trial_and_error}
\end{figure}

\begin{figure}[ht]
    \centering
    \includegraphics[width=1\linewidth]{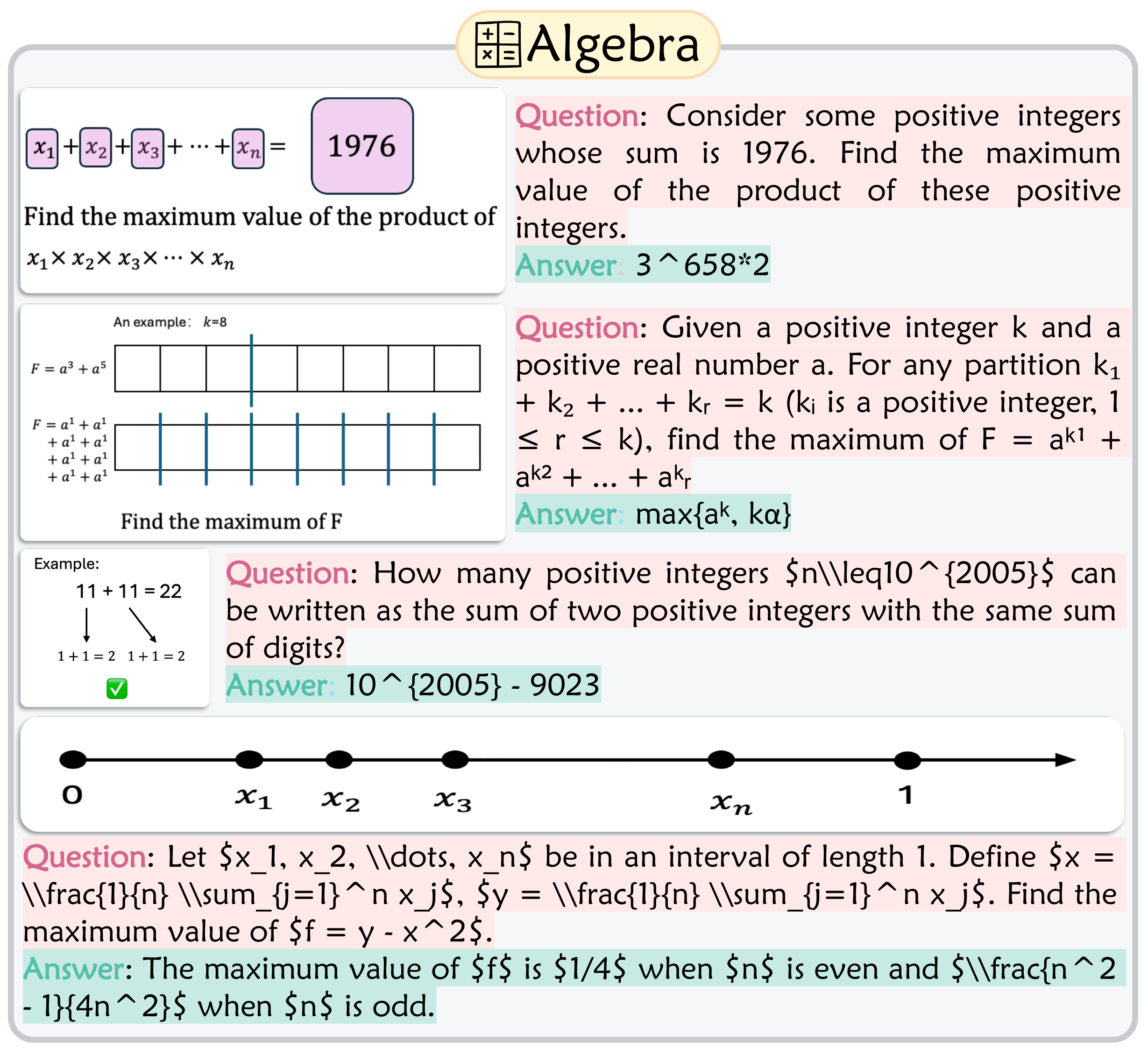}
    \caption{Examples of the algebra subcategory.}
    \label{fig:app_sub_algebra}
\end{figure}

\begin{figure}[ht]
    \centering
    \includegraphics[width=1\linewidth]{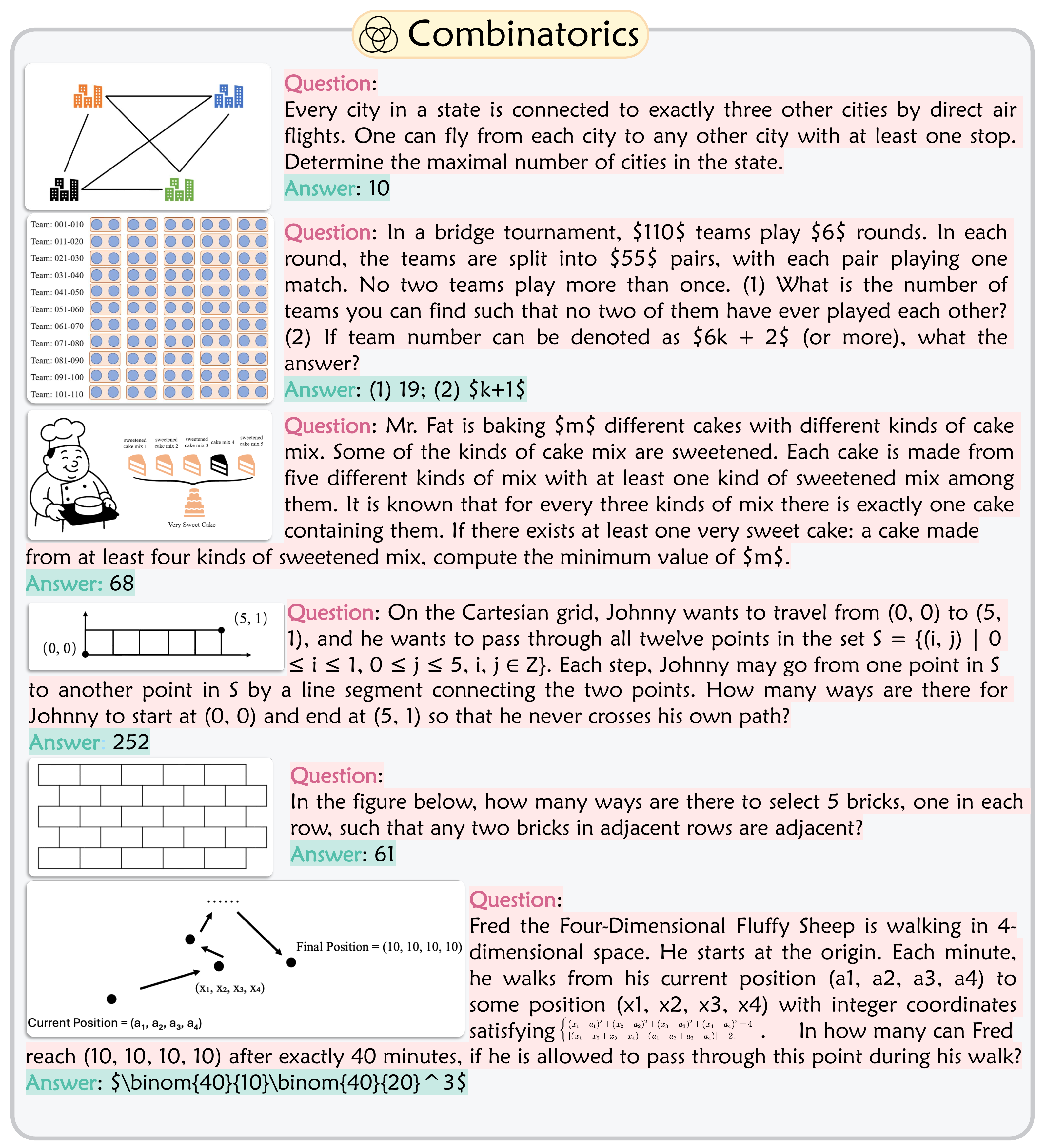}
    \caption{Examples of the combinatorics subcategory.}
    \label{fig:app_sub_combi}
\end{figure}

\begin{figure}[ht]
    \centering
    \includegraphics[width=1\linewidth]{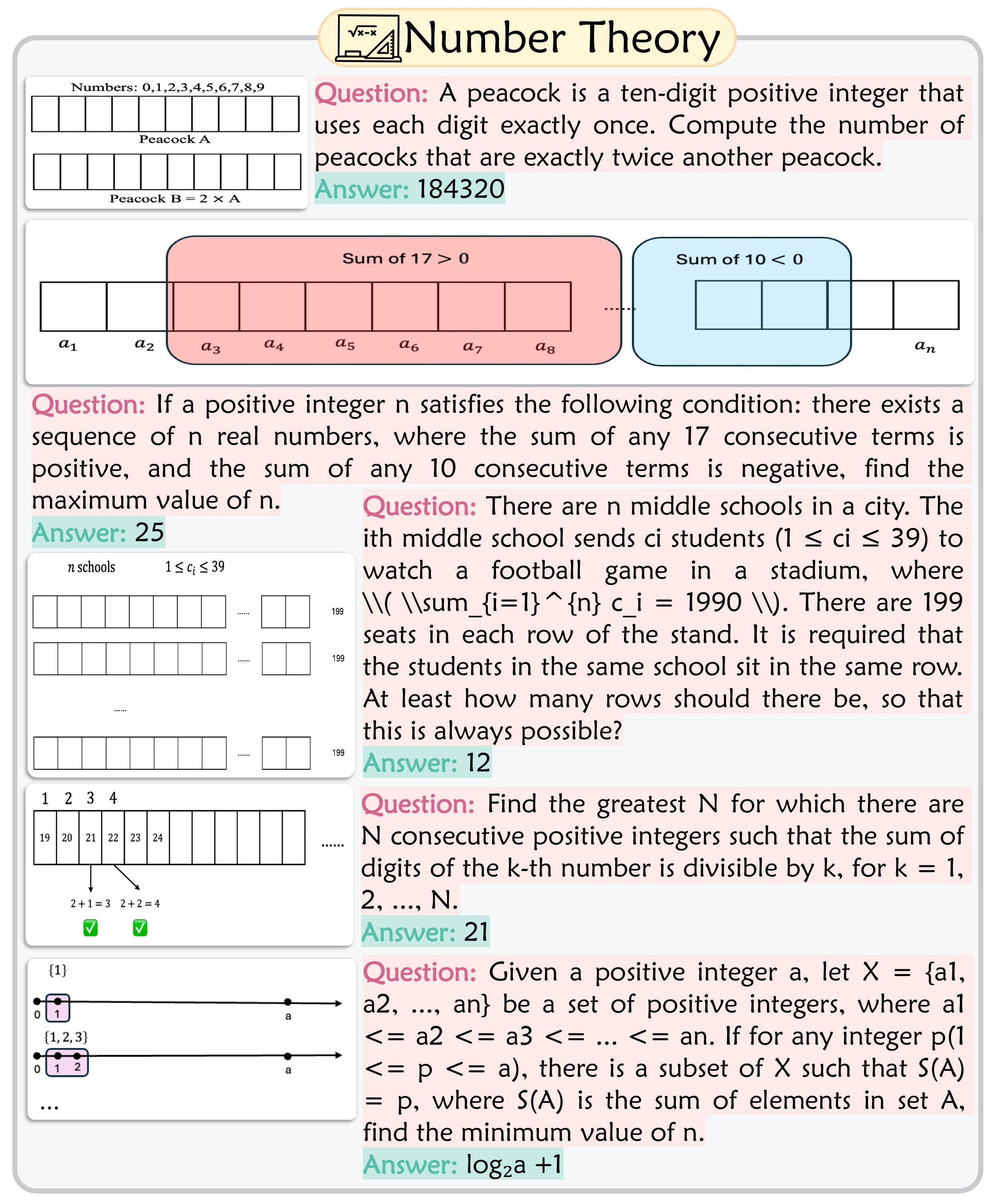}
    \caption{Examples of the number theory subcategory.}
    \label{fig:app_sub_number}
\end{figure}

\begin{figure}[ht]
    \centering
    \includegraphics[width=1\linewidth]{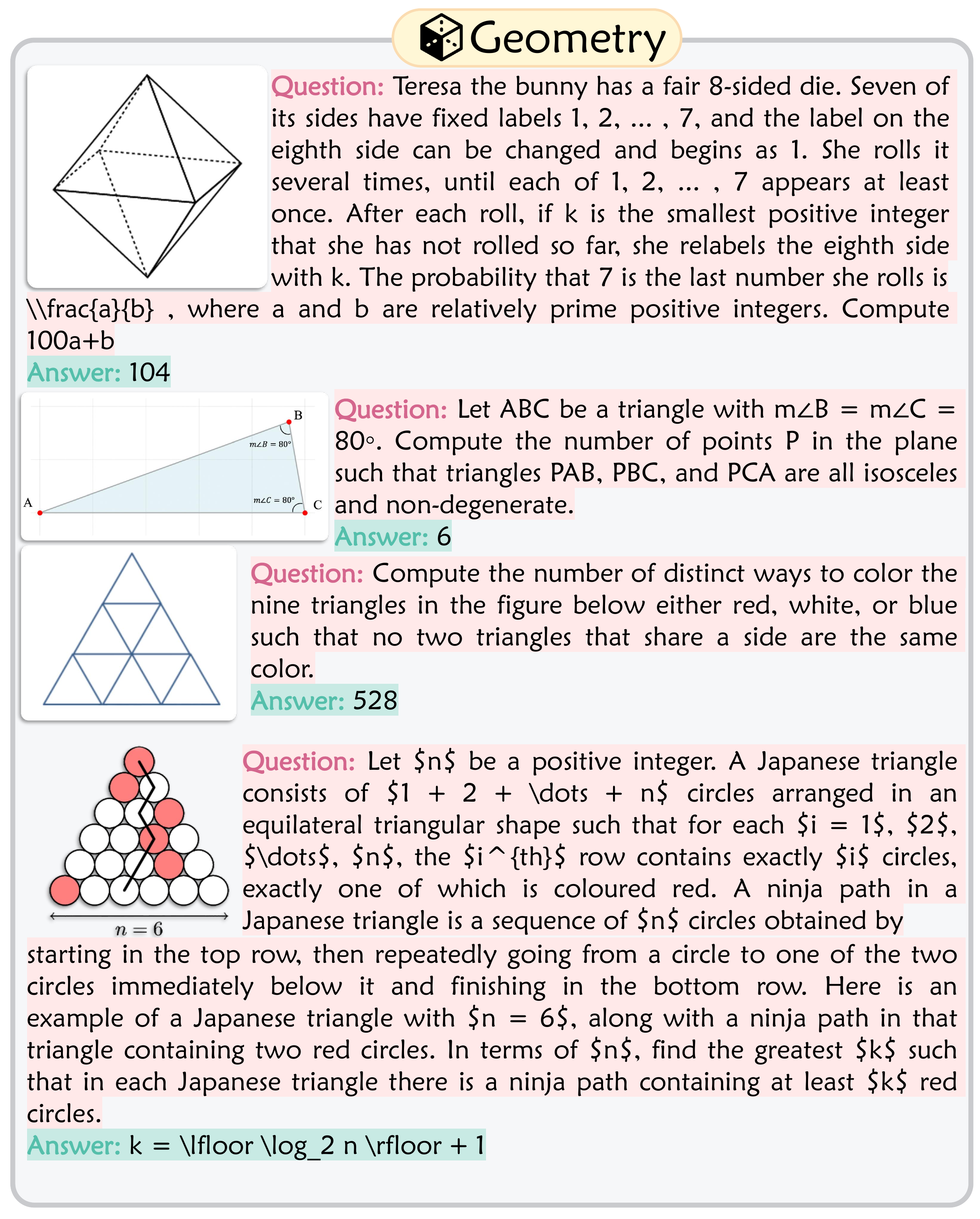}
    \caption{Examples of the geometry subcategory.}
    \label{fig:app_sub_geometry}
\end{figure}

\begin{figure}[ht]
    \centering
    \includegraphics[width=0.83\linewidth]{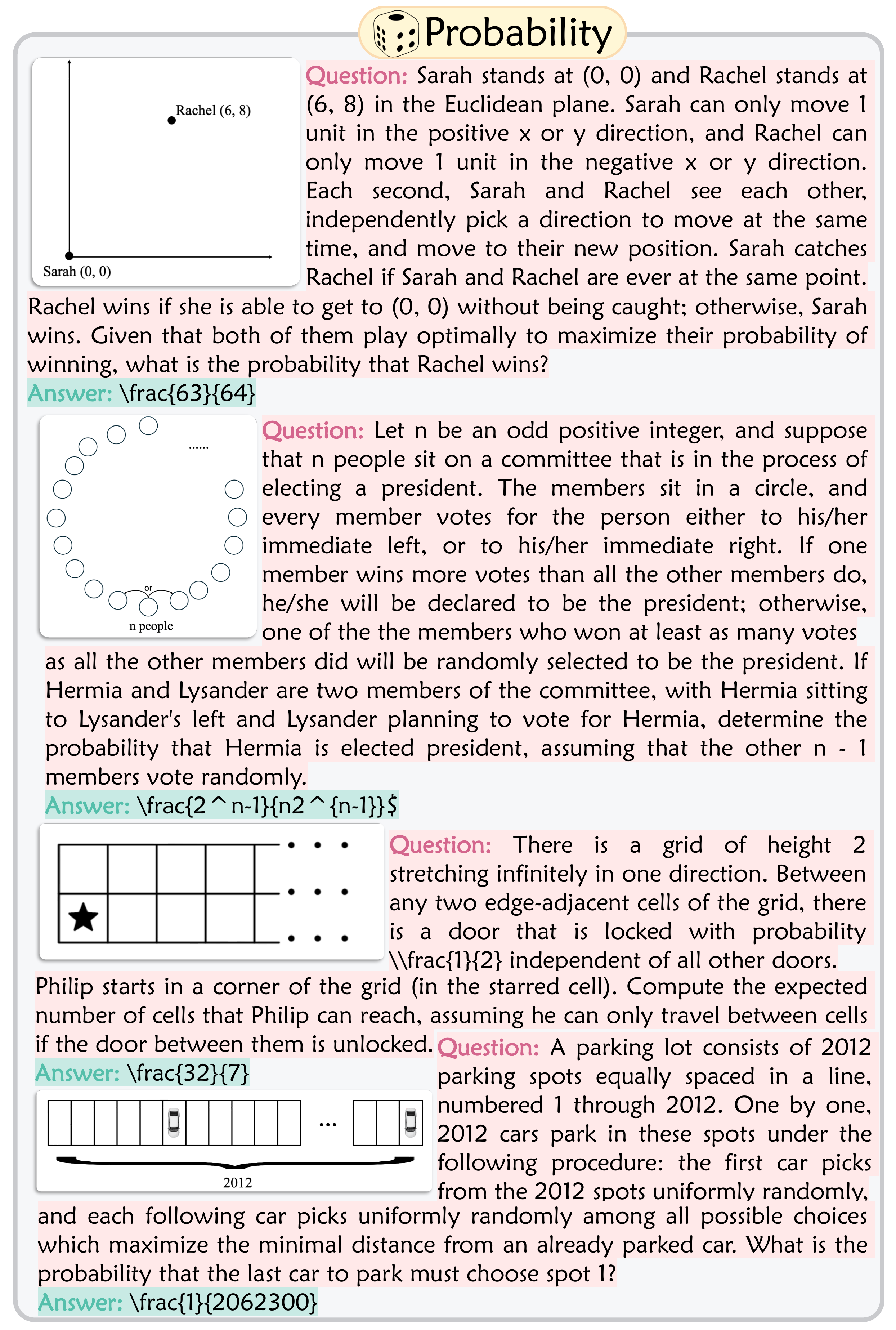}
    \caption{Examples of the probability subcategory.}
    \label{fig:app_sub_prob}
\end{figure}

\begin{figure}[ht]
    \centering
    \includegraphics[width=0.83\linewidth]{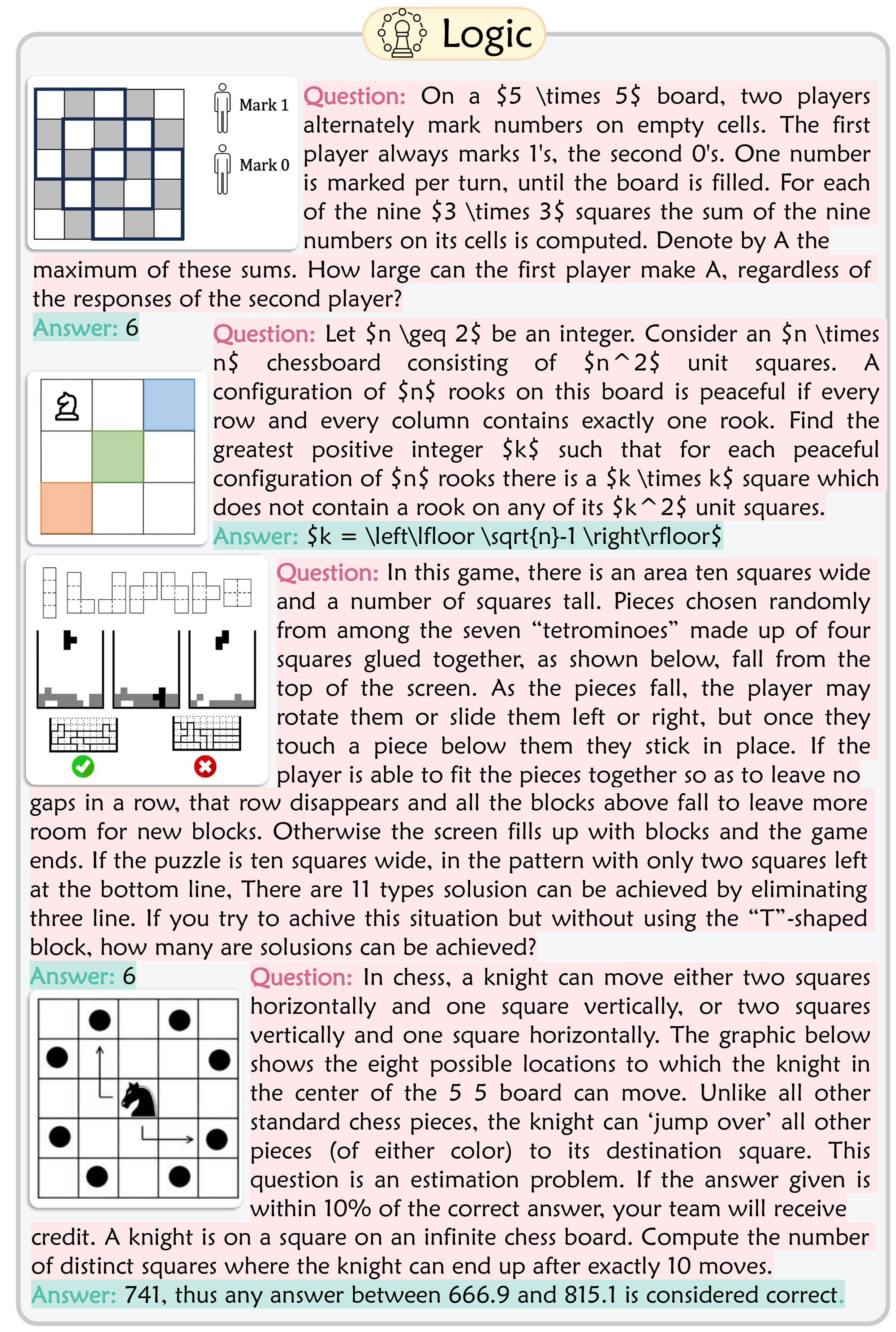}
    \caption{Examples of the logic subcategory.}
    \label{fig:app_sub_logic}
\end{figure}

\end{document}